\documentclass[lettersize,journal]{IEEEtran}
\usepackage{amsmath,amsfonts}
\usepackage{algorithmic}
\usepackage{algorithm}
\usepackage{array}
\usepackage[caption=false,font=normalsize,labelfont=sf,textfont=sf]{subfig}
\usepackage{textcomp}
\usepackage{stfloats}
\usepackage{url}
\usepackage{verbatim}
\usepackage{graphicx}
\usepackage{cite}
\usepackage{booktabs}
\usepackage{multirow}
\usepackage{xcolor}
\usepackage{hyperref}
\usepackage{threeparttable}

\begin{document}

\title{TSSM: Triaxial State Space Model for Global Station Weather Forecasting with Temporal-Variable-Historical Modeling}

\author{Songru Yang,
  Zili Liu$^\star$,
  Tao Han,
  Ben Fei,
  Fenghua Ling,
  Lei Bai,
  Chang Liu,
  Xiangyang Ji,
  Zhenwei Shi,~\IEEEmembership{Senior Member,~IEEE},
  Zhengxia Zou$^\star$,~\IEEEmembership{Senior Member,~IEEE}
  \thanks{
  This work was supported in part by the National Natural Science Foundation of China under Grants U24B20177, 62125102, U25A20401, and 62471014.}

  \thanks{
    Songru Yang, Zhenwei Shi and Zhengxia Zou are with the Department of Aerospace Intelligent Science and Technology, School of Astronautics, Beihang University, and with the Key Laboratory of Spacecraft Design Optimization and Dynamic Simulation Technologies, Ministry of Education, Beihang University, Beijing 100191, China.

    Zili Liu, Tao Han, Fenghua Ling, Lei Bai, are with Shanghai Artificial Intelligence Laboratory, Shanghai 200232, China.

    Ben Fei is with Department of Information Engineering, The Chinese University of Hong Kong, Hong Kong 999077, China

    Chang Liu, Xiangyang Ji are with Department of Automation, Tsinghua University, Beijing 100084, China
  }
}

\markboth{Journal of \LaTeX\ Class Files,~Vol.~14, No.~8, August~2021}%
{Shell \MakeLowercase{\textit{et al.}}: A Sample Article Using IEEEtran.cls for IEEE Journals}

\maketitle

\begin{abstract}
  Global Station Weather Forecasting (GSWF) is pivotal for localized and extreme weather prediction over key regions. Despite efforts to exploit look-back windows, existing methods show limited accuracy gains and struggle with extreme event and error accumulation. These limitations stem from over-reliance on short-term patterns, which are insufficient to capture chaotic weather dynamics, especially under partial observations. To address this problem, we propose a novel Triaxial State Space Model (TSSM) with a history-enhanced Temporal-Variable-Historical paradigm, which incorporates period-aligned historical weather data to compensate for long-term, large-scale periodic, and full-window weather patterns beyond the temporal look-back window. Specifically, TSSM stacks historical samples into period-aligned batches, where forecasting is causally supported by historical and current observations. Temporal, variable, and historical scanning are designed to capture axial temporal dependencies, variable correlations, and historical evolution. This structure is hierarchically shared to model seasonal to extreme events while alleviating misalignment across historical patterns. TSSM achieves SOTA performance on Weather-5K, the largest station weather dataset to date, with 10\% and 61\% gains in accuracy and extreme event metrics, and obtains 95\% best or second-best results on human-involved datasets. Its advantages are more pronounced in long-horizon and iterative forecasting, reaching a 37.5\% gain at 240h and up to 103.5\% under a 48h $\times$ 5 iterative setting. Moreover, TSSM retains $>$ 90\% performance under up to 80\% missing observations, compared with $<$ 43\% for baselines, demonstrating robustness and practical potential for reliable GSWF in global in-situ observation networks.
\end{abstract}


\section{Introduction}
\IEEEPARstart{G}{lobal} Station Weather Forecasting (GSWF) plays a vital role in delivering timely, localized weather predictions worldwide \cite{WMO2018GOS,wu2023interpretable,han2024weather}. By improving average accuracy, capturing extreme events, and extending effective forecasting horizons, GSWF supports critical applications such as disaster management and transportation through precise, site-specific forecasts. This capability is increasingly important for mitigating climate-change impacts and local-scale extremes that grid-based approaches often fail to resolve \cite{stott2016climate,zhu2026foundations}. Existing methods have already supported key scenarios such as the Winter Olympics \cite{wu2023interpretable}.

Existing data-driven approaches to GSWF generally formulate the task as multivariate time series forecasting. They use historical observations to predict future dynamics with advanced backbones, including MLP-based models \cite{zeng2023transformers,cyclenet,liu2024deriving}, LSTM variants \cite{karevan2020transductive,hewamalage2021recurrent}, Transformers \cite{zhou2022fedformer,liu2023itransformer,liu2022pyraformer}, and Mamba-based architectures \cite{ma2025timepro,gu2023mamba,liu2024mambads}. Recent studies have incorporated regional context \cite{huang2025regionalweather,liu2024deriving,liu2024mambads}. These methods aim to learn complex temporal-variable dependencies and nonlinear weather patterns, offering a compelling complement to traditional physics-based forecasting \cite{bauer2015quiet}. Nevertheless, even increasingly sophisticated designs \cite{wu2023interpretable,yang2025wssm} often provide limited gains, while average accuracy, extreme event prediction, and long-horizon stability remain unsatisfactory due to severe error accumulation in iterative forecasting.

We attribute these limitations to an overreliance on temporal look-back windows inherited from conventional time series forecasting. Under this paradigm, although previous methods build trend-seasonal-residual decomposition or retrieve memorized patterns \cite{cyclenet, wu2021autoformer}, they operate under stationarity and myopic assumptions \cite{box2015time,dickey1979distribution}, presuming that future patterns remain consistent with, and depend mainly on, a limited historical window. Such assumptions conflict with the inherently non-stationary, cross-scale periodic nature of atmospheric dynamics~\cite{milly2008stationarity,koscielny1998persistence}, thereby constraining the model's ability to capture long-range dependencies and the underlying physical structures essential for accurate forecasting.

\begin{figure}[!t]
  \centering
  \includegraphics[width=\linewidth]{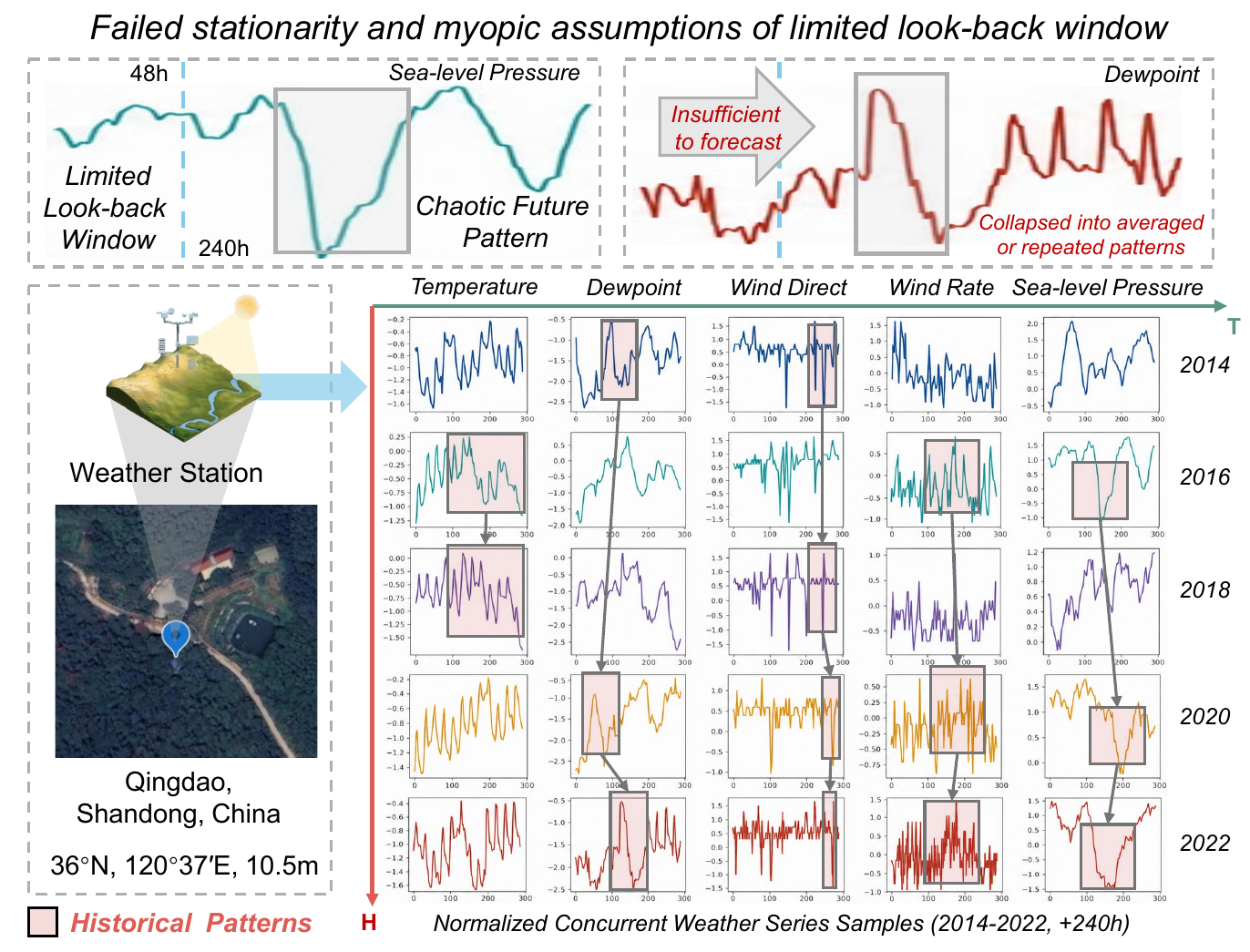}
  \caption{Consistent and separable historical weather patterns across aligned month/day/hour observations at a station in Qingdao, Shangdong, China. These shared patterns within each variable are highlighted by red boxes and gray arrows and tend to occur at historically aligned times. }
  \label{fig:intro1}
  \vspace{-5pt}
\end{figure}

Beyond the short-term temporal look-back window, we revisit GSWF from a long-term, cross-scale perspective. Inspired by same-period climatological analysis in traditional forecasting practice~\cite{lorenz1969analogues}, we observe that although weather series are highly chaotic along the current temporal dimension, they exhibit consistent and separable patterns in historical evolution, i.e., month/day/hour-aligned observations across years at the same station, as illustrated in Fig.~\ref{fig:intro1}.

\begin{figure*}[!t]
  \centering
  \includegraphics[width=\linewidth]{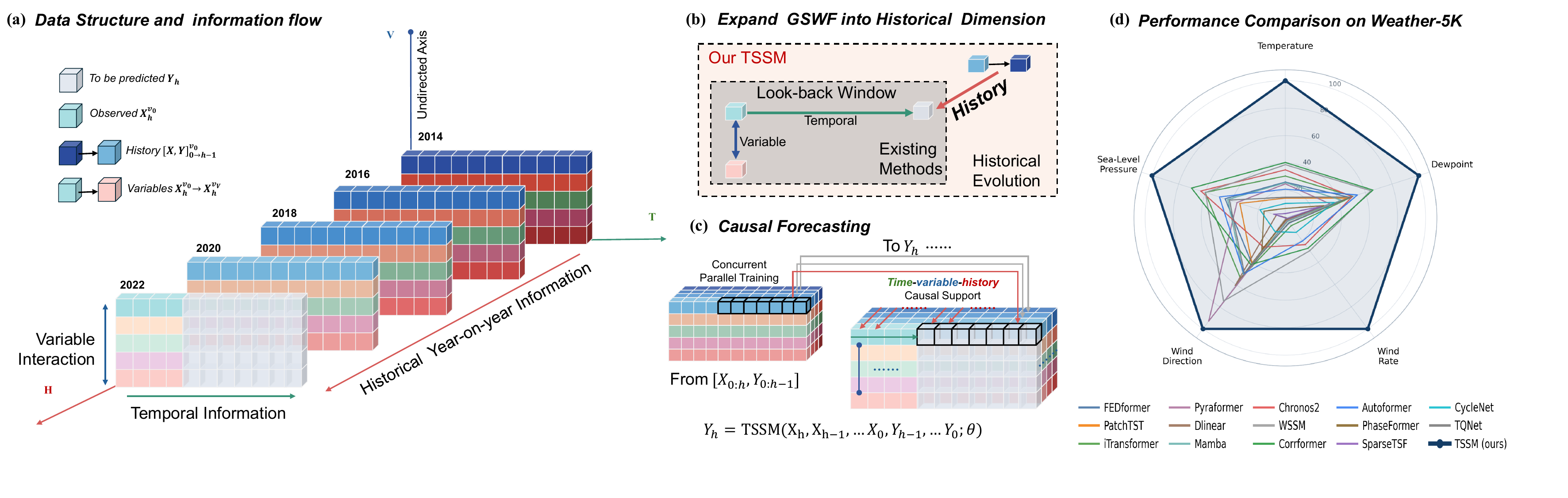}
  \caption{Overview of our extension of GSWF to the historical dimension through (a) data structure, (b) modeling strategy, (c) forecasting framework designed for the proposed Triaxial State Space Model, and (d) performance comparison on Weather-5K. Different methods are marked in colors}
  \label{fig:intro2}
\end{figure*}

Building on this insight, we expand temporal-variable modeling into a historical dimension and propose a novel Triaxial State Space Model (TSSM) with a temporal-variable-historical forecasting pipeline. As shown in Fig.~\ref{fig:intro2}, TSSM leverages the directed scanning capability of state space models for axial modeling, anchors short-term hourly dynamics on the temporal and variable axes to long-term yearly evolution through historical-axis modeling, and causally learns evolutionary patterns from the same month, day, and hour across historical years. This design provides long-term references for forecasting, improves extreme event capture especially for long-horizon settings, and calibrates predictions to mitigate accumulated errors in iterative forecasting. Moreover, TSSM extends the existing GSWF setting to support forecasting under missing observations \cite{park2019temperature}, thereby improving robustness in real-world scenarios where missing data are unavoidable.

Specifically, our pipeline contains three key designs: \textbf{triaxial data reorganization}, \textbf{causal forecasting}, and the \textbf{TSSM architecture}. First, for \textbf{triaxial data reorganization}, we establish a historical axis without extra inputs by stacking multivariate weather series from previous years at the same month/day/hour into a $T\times V\times H$ tensor, corresponding to temporal, variable, and historical axes, as shown in Fig.~\ref{fig:intro2}(a). Second, for \textbf{causal forecasting}, each current prediction is supported by historical windows from years $0$ to $H$ in the look-back window and years $0$ to $H-1$ in the forecast window, while temporal-variable patterns flow causally from look-back to forecast windows, no future leakage as Fig.~\ref{fig:intro2}(c). Third, for the \textbf{TSSM architecture}, we introduce temporal, variable, and historical scanning modules, i.e., T-Scan, V-Scan, and H-Scan, to efficiently model yearly expanded inputs. T-Scan captures short-term dynamics, V-Scan models inter-variable correlations, and H-Scan captures long-term historical evolution and deviations from historical averages toward recent years, as shown in Fig.~\ref{fig:intro2}(b). These modules can be independently combined and hierarchically shared to construct weather states from seasonal patterns to extremes. When available history decreases, TSSM naturally degenerates to temporal-variable prediction anchored by last-year references, and shorter cycles such as weekly aligned data can also be adopted.

We conduct extensive experiments on Weather-5K \cite{han2024weather5k}, our main testbed, currently the largest GSWF-specific station weather dataset that covers and extends existing station datasets, and further validate generalizability on hydrological data under human regulation and general time series datasets spanning human behavioral patterns and energy dynamics \cite{li2025mcann, zhou2021informer, lai2018lstnet}. It contains hourly temperature, dew point, wind direction, wind rate, and sea-level pressure records from 5k+ surface weather stations worldwide, spanning six continents over a 10-year period. In both qualitative and quantitative evaluations, TSSM significantly outperforms existing methods in terms of both average forecasting accuracy and extreme event prediction accuracy as shown in Fig.~\ref{fig:intro2}(d). For Weather-5K, our method achieves the best or second-best performance on 88\% of the 72 results across four metrics, with average gains of 10\% in overall accuracy and 61\% in extreme event prediction. For other 2 datasets, TSSM also achieves in total 95\% best or second-best results. Moreover, our advantages become even more pronounced in long-term and robust forecasting. It reaches a 37.5\% overall gain at 240h and up to 103.5\% under 48h $\times$ 5 iterative manner with cumulative iterative error $<$4.3\%, even outperforming one-step forecasting, and exhibits strong robustness to missing data, with less than 10\% degradation compared with $>$ 50\% for baselines, at an 80\% missing rate.

Our contributions can be summarized as follows:

\label{sec:contributions}
\begin{itemize}
  \item We introduce TSSM that extends GSWF to a triaxial temporal-variable-historical paradigm for jointly modeling short-term dynamics and long-term evolution while compensating for full-window stationarity and cross-scale periodicity beyond look-back windows, offering a nearly free lunch for comprehensive performance gains.
  \item We propose triaxial scanning mechanisms, including T-Scan, V-Scan, and H-Scan. Based on short-term dynamics and variable correlations, we extend historical-axis modeling in an isomorphic state space model manner to learn the long-term evolution patterns of period-aligned historical weather and design a history-enhanced pipeline with triaxial data reorganization and causal forecasting.
  \item Extensive experiments on Weather-5K, the largest up-to-date GSWF dataset and human-involved datasets, validate that TSSM not only significantly outperforms existing methods in average and extreme event prediction but also provides an unprecedented solution to cumulative errors in iterative forecasting and missing observations without extra information or high complexity.

\end{itemize}

\section{Related Works}
\subsection{Data-driven Weather Prediction}
Since 2022, the AI and atmospheric science communities have shown rapidly growing interest in data-driven Numerical Weather Prediction (NWP) models, such as Pangu-Weather~\cite{bi2023pangu}, GraphCast~\cite{lam2023graphcast}, GenCast~\cite{price2024gencast}, and related methods~\cite{pathak2022fourcastnet,chen2023fuxi,espeholt2022metnet3}, which operate on gridded reanalysis data (e.g., ERA5~\cite{hersbach2020era5}) at resolutions such as $0.25^\circ$ and $0.09^\circ$. In gridded weather-field forecasting and nowcasting, recent studies have emphasized precipitation/rainfall dynamics and multi-source fusion \cite{wang2023radarheavyrain,li2024shortlongprecip,yu2024coordrainfall,chen2024cloud3dnowcasting,liu2025pwvradar,zhang2025rsggan,liu2025weightedloss}. However, these large-scale, region-averaged methods may not match station-level weather systems and can overlook highly destructive local extreme weather. In contrast to weather-field forecasting, station-level forecasting focuses on point-wise in-situ observations and localized physical regimes, some initial attempts, such as Corrformer and WSSM~\cite{wu2023interpretable, yang2025wssm,huang2025regionalweather}, have treated station weather forecasting as an independent task and achieved promising results. However, these methods still rely on short-term temporal look-back windows, which greatly limits their capability, robustness and real-world applicability.

\subsection{Multivariate Time Series Forecasting}
Multivariate Time Series Forecasting aims to predict future dynamics from short-term temporal-variable dependencies within a look-back window, implicitly assuming local stationarity and dominant dependence on recent history~\cite{zeng2023transformers,nie2023patchtst,wu2023timesnet,cyclenet,gao2022earthformer,han2024softs,liu2022nonstationary,liu2024timeffm,ma2025neuripsmofo,nguyen2023climax,ravuri2021skilful,wang2024news2forecast,wu2021autoformer,wu2023corrformer,zhang2023crossformer,zheng2024correlated}. Early explorations mainly relied on CNN/RNN backbones to extract local temporal patterns, while simple pooling-like operations were used to model variable correlations~\cite{karevan2020transductive,hewamalage2021recurrent}. More recently, Transformer-based methods have enabled more flexible factorized modeling of time-variable interactions via attention mechanisms~\cite{zhou2022fedformer,liu2023itransformer,liu2022pyraformer,das2023timesfm,ansari2024chronos}. In parallel, SSM-based methods with directional scanning have improved long-range dependency modeling and computational efficiency~\cite{gu2022s4,gu2023mamba,liu2024mambads,ma2025timepro}. A recent shared trend across both Transformer and SSM methods is the shift toward time series foundation models and LLM-enhanced pipelines, aiming for better transferability, robustness, and long-horizon stability~\cite{das2023timesfm,ansari2024chronos,dao2024transformersssm}. Existing GSWF methods inherit this framework and have achieved promising results~\cite{wu2023interpretable,yang2025wssm}. However, although incorporating geographical information or multi-station modeling can bring marginal improvements, forecasting solely from a look-back window is inherently limited in capturing long-term weather structures, such as annual or seasonal backgrounds. These methods also lack physical structure~\cite{milly2008stationarity,koscielny1998persistence}: even a simple historical average can provide significant stationarity support for understanding extreme weather. As a result, predictions often degenerate into repetitions of look-back-window patterns, largely due to stationarity and myopic assumptions. This motivates our preliminary extension of GSWF beyond the look-back window into a historical dimension.

\subsection{Period-aware and Memory-enhanced Forecasting}
Periodicity has been extensively studied in time series forecasting, as it provides a useful prior for modeling long-range temporal dependencies in a simplified and robust manner. Existing methods, such as~\cite{wu2021autoformer,niu2026phaseformer,yuan2024diffusionts}, usually establish cross-period dependencies by associating positions with the same phase across different cycles. Nevertheless, unlike our method, these approaches still extract periodic patterns only from the limited look-back window. When the observation window is insufficient or corrupted, such locally mined periodic features can become unreliable, leading to degraded performance.

Taking a step further, memory mechanisms are an effective strategy for improving long-horizon forecasting performance, but they have only been sparsely explored in time series forecasting. Existing methods typically treat periodicity as a mechanism for memory pattern extraction, where sequences are segmented based on periods and representative patterns are stored to enhance forecasting, such as ~\cite{cyclenet,lin2025tqnet}. More recent work either augments recurrent models with long-memory filters~\cite{yang2024mgru,krausxlstm} or adopts retrieval-augmented LLMs for memory-enhanced forecasting~\cite{tao2026memcast}. Compared with our method, however, these approaches typically require explicit memory-bank design and maintenance, they often merge or select memory items using statistical criteria, such as feature similarity or attention, which may discard physically meaningful structures and cause pattern mismatch, such as injecting winter memories into summer forecasts.

\section{Method}
\begin{figure*}[!t]
  \centering
  \includegraphics[width=\linewidth]{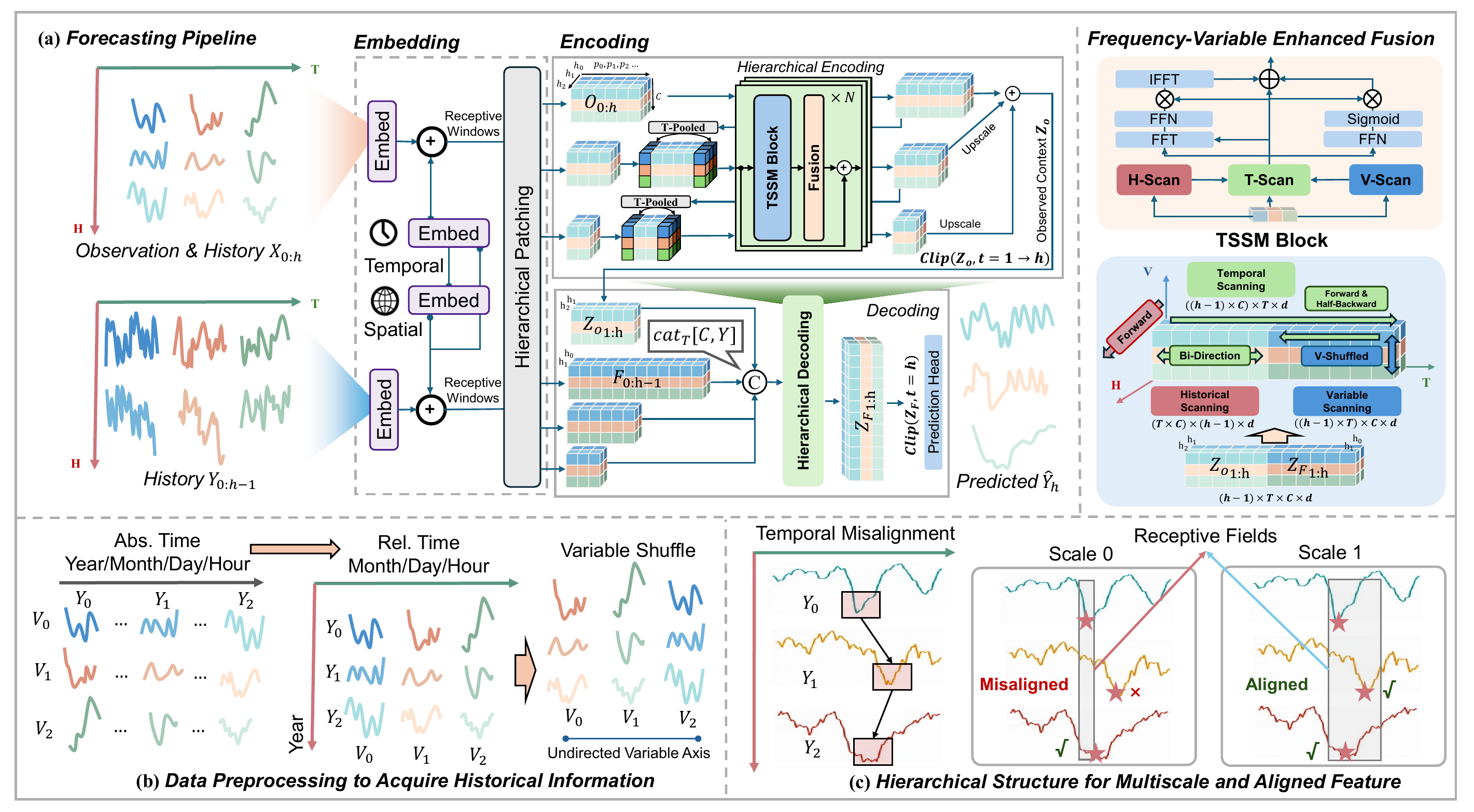}
  \caption{Detailed architecture of the TSSM including (a) model architecture and forecasting pipeline, (b) triaxial reorganized data preprocessing, and (c) design motivation and function of the hierarchical structure.}
  \label{fig:method}
\end{figure*}

An overview of the Triaxial State Space Model (TSSM) is shown in Fig.~\ref{fig:method}. We pioneer the modeling of historical evolution as a state-transition process along the historical axis, describe coexisting short-long-term weather dynamics in an isomorphic SSM manner, and maintain axial linear complexity under multiplied data dimensions. TSSM extends the GSWF pipeline through two co-designed components: the triaxially reorganized data structure and the causal forecasting strategy.

\subsection{Triaxial Reorganized Data}
Incorporating a historical axis requires reorganizing conventional time-variable 2D data into a time-variable-history 3D representation, as shown in Fig.~\ref{fig:intro2}(a). Rather than introducing additional external inputs, we extract intrinsic historical context through period-aligned analysis, as illustrated in Fig.~\ref{fig:method}(b), multivariate records $V_{i}$ from different years $Y_{i}$ in absolute time are stacked along the historical dimension on a relative temporal axis aligned by the month, day, and hour, which can be readily integrated into existing pipelines.

Let $\mathbf{X}_{\text{all}} \in \mathbb{R}^{T_{\text{all}} \times V}$ denote the complete weather series of a station, where $T_{\text{all}}$ is the total number of time steps and $V$ is the number of variables. Through period-aligned alignment, the temporal axis is folded so that samples with identical sub-timestamps (i.e., the same month/day/hour) are aligned across $H$ years. This reshaping yields a historically augmented tensor $\mathbf{X}_{\text{ch}} \in \mathbb{R}^{H \times T_{\text{all}}/H \times V}$. To generate the final triaxial input $\mathbf{X} \in \mathbb{R}^{H \times T \times V}$, we apply a sliding window of length $T$ along the intra-year time axis. Among the three axes, the temporal and historical axes are ordered, whereas the variable axis is unordered; therefore, we randomly shuffle the variable order for each sample. The construction is defined as:

\begin{equation}
  (\mathbf{X})_{h, t, v} = (\mathbf{X}_{\text{all}})_{h \cdot \frac{T_{\text{all}}}{H} + i + t, \pi(v)}
  \label{eq:combined_process}
\end{equation}

where $h \in \{0, 1, \dots, H-1\}$ denotes the year index, $t \in \{0, 1, \dots, T-1\}$ denotes the time index within the sliding window, $i$ denotes the start index along the intra-year time axis, and $\pi(\cdot)$ is a permutation over variable indices.

\subsection{Causal Forecasting}
Introducing a historical dimension redefines the forecasting task by requiring both historical and temporal causality to be modeled. To predict the future window of the $h$-th year, we use the look-back windows from years $0$ to $h$ and the future windows from years $0$ to $h-1$ as conditions, as briefly shown in Fig.~\ref{fig:intro2}(c) and illustrated across the full pipeline in Fig.~\ref{fig:method}(a).

Consequently, information flows unidirectionally along two axes: historically, look-back information propagates from year $0$ to $h$, while period-aligned history in the future window propagates from year $0$ to $h-1$; temporally, information propagates from the look-back window to the future, as in existing GSWF settings. This dual-causality mechanism has a two-fold interpretation: long-term historical evolution is dynamically calibrated by short-term patterns within the look-back window, while these short-term patterns are constrained by the long-term background. In fact, this long--short-term correlation reflects multi-periodic weather processes.

Samples from all years $h \in [0, H)$ are processed simultaneously during training, thereby maintaining overall computational efficiency. The index of current year is $H-1$ The computation and optimization are formulated as follows:

\begin{align}
  \hat{\mathbf{Y}}_h &= \mathcal{F}_{\Theta}(\mathbf{X}_{0:h}, \mathbf{Y}_{0:h-1})\\
  \mathcal{L}_{\text{MSE}} &= \frac{1}{H} \sum_{h=0}^{H-1} \left\| \hat{\mathbf{Y}}_h - \mathbf{Y}_h \right\|_F^2
  \label{eq:dual_causality}
\end{align}

where $\hat{\mathbf{Y}}_h$ denotes the predicted future for the $h$-th year, and $\mathcal{F}_{\Theta}$ denotes TSSM parameterized by $\Theta$ and optimized with $\mathcal{L}_{\text{MSE}}$. The term $\mathbf{X}_{0:h} = \{\mathbf{X}_0, \dots, \mathbf{X}_{h-1}\}$ represents look-back windows from years $0$ to $h$. The term $\mathbf{Y}_{0:h-1} = \{\mathbf{Y}_0, \dots, \mathbf{Y}_{h-2}\}$ represents future windows from preceding years. During inference, the forecast $\hat{\mathbf{Y}}_{H-1}$ is causally conditioned on full observations $\mathbf{X}_{0:H}$ and $\mathbf{Y}_{0:H-1}$.

\subsection{Triaxial State Space Model}
With historically stacked data, modeling global correlations leads to a dramatic increase in complexity. Our Triaxial State Space Model efficiently fuses temporal, variable, and historical information through dedicated scanning branches along each axis, leveraging the directed scanning mechanism of state space models. These branches can be fused independently and shared hierarchically to construct weather-state representations ranging from seasonal patterns to extreme events. The detailed design of TSSM and its operation within the forecasting pipeline are shown in Fig.~\ref{fig:method}(a).

For clarity, we present the mechanisms in the order of historical scanning (H-Scan), variable scanning (V-Scan), and temporal scanning (T-Scan). We denote the observation encoding of $\mathbf{X}_{0:H}$ within the look-back window $T_o$ as $\mathbf{O}_{0:H}$ and the historical future encoding of $\mathbf{Y}_{0:H-1}$ as $\mathbf{F}_{0:H-1}$. Both are processed by a time-step-wise $\text{Tokenizer}(\cdot)$ with latent dimension $d$ as follows:

\vspace{-10pt}
\begin{equation}
  \begin{aligned}
    \mathbf{O}_{0:H} &= \text{Tokenizer}(\mathbf{X}_{0:H}) \in \mathbb{R}^{(H) \times T_o \times C \times d},\\
    \mathbf{F}_{0:H-1} &= \text{Tokenizer}(\mathbf{Y}_{0:H-1}) \in \mathbb{R}^{(H-1) \times (T-T_o) \times C \times d}\\
  \end{aligned}
\end{equation}

\subsubsection{Historical Scanning}
Along the historical axis, the historical scanning branch constructs an anomaly-enhanced unidirectional SSM starting from year $0$. After reshaping the features into $\mathbb{R}^{(T \times C) \times H \times d}$ to explicitly decouple the historical axis, we first scan historical values to extract evolution patterns $\mathbf{f}_{o}^{h}$ and $\mathbf{f}_{f}^{h}$. The computations are as follows:

\begin{equation}
  \begin{aligned}
    \mathbf{O}_{0:H} &\xrightarrow{\text{reshape}} \mathbb{R}^{(T_o \times C) \times (H) \times d}, \\
    \mathbf{F}_{0:H-1} &\xrightarrow{\text{reshape}} \mathbb{R}^{((T-T_o) \times C) \times (H-1) \times d}, \\
    \mathbf{f}_{o}^{h} &= \text{SSM}(\mathbf{O}_{0:H}), \\
    \mathbf{f}_{f}^{h} &= \text{SSM}(\mathbf{F}_{0:H-1})
  \end{aligned}
\end{equation}

More importantly, we scan the deviations between each historical value and the their average to uncover extreme patterns $\mathbf{f}_{o}^{e}$ and $\mathbf{f}_{f}^{e}$. This operation is directly motivated by how climatological analysis evaluates extreme weather: anomaly values that deviate significantly from the historical average $\text{Avg-h}(\cdot)$ indicate potential extreme events. The evolution and extreme patterns are then fused by MLPs to obtain historical features $\mathbf{Z}_O^H$ and $\mathbf{Z}_F^H$. The computations are as follows:

\begin{equation}
  \begin{aligned}
    \mathbf{f}_{o}^{e} &= \text{SSM}(\left|\mathbf{O}_{0:H}-\text{Avg-h}(\mathbf{O}_{0:H})\right|), \\
    \mathbf{f}_{f}^{e} &= \text{SSM}(\left|\mathbf{F}_{0:H-1}-\text{Avg-h}(\mathbf{F}_{0:H-1})\right|), \\
    \mathbf{Z}_O^H, \mathbf{Z}_F^H &= \text{MLP}(\mathbf{f}_{o}^{h} + \mathbf{f}_{o}^{e}), \text{MLP}(\mathbf{f}_{f}^{h} + \mathbf{f}_{f}^{e})
  \end{aligned}
\end{equation}

\subsubsection{Variable Scanning}
Along the variable axis, because unordered modeling has already been handled during data reorganization, the variable scanning branch adopts a simple SSM. After reshaping into $\mathbb{R}^{(T \times H) \times C \times d}$ to explicitly decouple the variable axis, we capture inter-variable correlations $\mathbf{Z}_{O}^V$ and $\mathbf{Z}_{F}^V$. Computations are as follows:

\vspace{-10pt}
\begin{equation}
  \begin{aligned}
    & \mathbf{O}_{0:H} \xrightarrow{\text{reshape}} \mathbb{R}^{(T_o \times H) \times C \times d}, \\
    & \mathbf{F}_{0:H-1} \xrightarrow{\text{reshape}} \mathbb{R}^{((T-T_o) \times (H-1)) \times C \times d}, \\
    & \mathbf{Z}_{O}^V = \text{MLP}(\text{SSM}(\mathbf{O}_{0:H})),\\
    & \mathbf{Z}_{F}^V = \text{MLP}(\text{SSM}(\mathbf{F}_{0:H-1}))\\
  \end{aligned}
  \label{eq:variable_scanning}
\end{equation}

Notably, shuffling the variable order forces the model to distinguish and interact with different variables based on their content rather than their relative positions along the axis.

\subsubsection{Temporal Scanning}
Along the temporal axis, the temporal scanning branch fuses features from the other branches and generates the final prediction. It includes a bidirectional SSM to extract temporal dependencies in the look-back window and a forward and half-backward SSM to model temporal dependencies from preceding years while injecting triaxial look-back conditions into the current future window. The inputs are reshaped into $\mathbb{R}^{(H \times C) \times T \times d}$ to explicitly decouple the temporal axis.

For $\mathbf{O}_{0:h}$ in the look-back window, we perform bidirectional scanning to extract the temporal dependency $\mathbf{Z}_{O}^T$. Together with the historical feature $\mathbf{Z}_O^H$ and variable correlation $\mathbf{Z}_{O}^V$, it forms the triaxial look-back condition $\mathbf{Z}_{O}$. The computations are as follows:

\vspace{-10pt}
\begin{equation}
  \begin{aligned}
    &\mathbf{O}_{0:H} \xrightarrow{\text{reshape}} \mathbb{R}^{(H \times C) \times T_o \times d}, \\
    &\mathbf{Z}_{O}^T = \text{MLP}(\text{SSM}(\mathbf{O}_{0:H}) + \text{flip}(\text{SSM}(\text{flip}(\mathbf{O}_{0:H})))), \\
    &\mathbf{Z}_O^H, \mathbf{Z}_O^V \xrightarrow{\text{reshape}} \mathbb{R}^{(H \times C) \times T_o \times d}, \\
    & \mathbf{Z}_{O} = \text{Fus}(\mathbf{Z}_{O}^T + \mathbf{Z}_O^H + \mathbf{Z}_{O}^V) \in \mathbb{R}^{(H \times C) \times T_o \times d}, \\
    &\mathbf{Z}_{O} \xrightarrow{\text{Clip 1:H}} \mathbb{R}^{((H-1) \times C) \times T_o \times d}, \\
  \end{aligned}
  \label{eq:temporal_scanning}
\end{equation}

For $\mathbf{F}_{0:H-1}$ in the future window, we first concatenate $\mathbf{Z}_{O}$ with $\mathbf{F}_{0:H-1}$, then perform a forward scan to inject conditions unidirectionally. Its output $\mathbf{f}_{f}^t$ is fused with a half-backward scan output $\mathbf{f}_{b}^t$ limited to the future window to generate the future temporal dependency $\mathbf{Z}_{F}^T$ conditioned on look-back observations. Computations are as follows:

\vspace{-10pt}
\begin{equation}
  \begin{aligned}
    &\mathbf{F}_{0:H-1} \xrightarrow{\text{reshape}} \mathbb{R}^{((H-1) \times C) \times (T-T_o) \times d}, \\
    & \mathbf{f}_{f}^t = \text{SSM}(\text{Concat}[\mathbf{Z}_{O}, \mathbf{F}_{0:H-1}]) \in \mathbb{R}^{((H-1) \times C) \times T \times d}, \\
    & \mathbf{f}_{b}^t = \text{flip}(\text{SSM}(\text{flip}(\mathbf{F}_{0:H-1}))), \\
    & \mathbf{Z}_{F}^T = \text{MLP}(\mathbf{f}_{f}^t[:,T_o:,:]+\mathbf{f}_{b}^t)
  \end{aligned}
  \label{eq:temporal_scanning}
\end{equation}

By combining the historical feature $\mathbf{Z}_F^H$ and variable correlation $\mathbf{Z}_{F}^V$, we generate the triaxial future latent $\mathbf{Z}_{F}$ and the final prediction for the current year, $\hat{\mathbf{Y}}_{H-1}$, as follows:

\vspace{-10pt}
\begin{equation}
  \begin{aligned}
    &\mathbf{Z}_F^H, \mathbf{Z}_F^V \xrightarrow{\text{reshape}} \mathbb{R}^{((H-1) \times C) \times (T-T_o) \times d}, \\
    & \mathbf{Z}_{F} = \text{Fus}(\mathbf{Z}_{F}^T + \mathbf{Z}_F^H + \mathbf{Z}_{F}^V)\in \mathbb{R}^{((H-1) \times C) \times (T-T_o)\times d}, \\
    & \hat{\mathbf{Y}}_{1:H} = \text{MLP}(\mathbf{Z}_{F})\xrightarrow{\text{reshape}} \mathbb{R}^{(H-1) \times (T-T_o) \times C}, \\
    &\hat{\mathbf{Y}}_{H-1} = \mathbf{Y}_{1:H}[-1, :, :] \in \mathbb{R}^{(T-T_o) \times C}
  \end{aligned}
  \label{eq:temporal_scanning}
\end{equation}

The fusion module $\text{Fus}(\cdot)$ above employs frequency and variable enhancement. Given input $\mathbf{u}\in \mathbb{R}^{M \times N \times d}$ with sequence length $N$ and channel size $M$, learnable filters $\mathbf{F}$ and MLP modulators $\text{M}(\cdot)$ are used to adaptively select crucial frequency components and identify variable-specific patterns. We denote the corresponding frequency- and variable-enhanced latents as $\mathbf{u}_f$ and $\mathbf{u}_v$. Computations are as follows:

\begin{equation}
  \label{TFM}
  \begin{aligned}
    \mathbf{u} & = \text{MLP}(\mathbf{u}),\\
    \mathbf{u}_f &= \text{iFFT}(\text{FFT}(\mathbf{u})\cdot\text{FFT}(F)), \\
    \mathbf{u}_v &= \mathbf{u}_f\cdot \text{M}(\text{Pool}(\mathbf{u}_f))\\
  \end{aligned}
\end{equation}

\subsubsection{Hierarchically Shared Structure}
We organize TSSM into a hierarchical structure to process weather-series features at multiple temporal resolutions. This design is motivated by two considerations: (1) weather series exhibit multi-scale temporal dynamics, and hierarchical processing enables coarse-to-fine representations ranging from global periodic patterns to local extreme events; and (2) although lie on the same temporal axis, patterns from different years are difficult to align precisely in time. Hierarchical processing therefore expands the historical receptive field and better captures shared patterns in historical evolution.

As shown in Fig.~\ref{fig:method}(b), we first perform hierarchical patching on the weather series to obtain a set of micro-to-macro features, denoted as [$\mathbf{p}_{N_0}$, $\mathbf{p}_{N_1}$, \ldots, $\mathbf{p}_{N_l}$], where $N_i$ is the number of patches. We then use TSSM to process these features sequentially from $N_l$ to $N_0$. Between two successive scales, we insert average-pooled features from the previous scale at the beginning and end of the current sequence, enabling the encoder to explicitly exploit the previous scale's global prior when processing the current scale. Across all scales, we bilinearly upsample lower-resolution features to the highest resolution to form the final hierarchically encoded output $\mathbf{z}$. Given the input embedding $\mathbf{u}\in \mathbb{R}^{M \times N \times d}$, the computations are as follows:
\begin{equation}
  \begin{aligned}
    &\mathbf{p}_{N_i} = \text{Patch}(\mathbf{u}, \text{patchsize}_i)\in\mathbb{R}^{M \times N_i \times d},\\
    &\mathbf{z}_{N_i} = \text{TSSM}(\mathbf{p}_{N_i}), \\
    &\mathbf{hp}_{N_i} = \text{Avg}(\mathbf{z}_{N_i})\in\mathbb{R}^{M \times 1 \times d},\\
    &\mathbf{p}_{N_{i-1}} = \text{Patch}(\mathbf{u}, \text{patchsize}_{i-1})\in\mathbb{R}^{M \times N_{i-1} \times d},\\
    &\mathbf{p}_{N_{i-1}} = \text{Concat}[\mathbf{hp}_{N_i}, \mathbf{p}_{N_{i-1}}, \mathbf{hp}_{N_i}],\\
    &\mathbf{z}_{N_{i-1}} = \text{TSSM}(\mathbf{p}_{N_{i-1}})[1:-1], \\
    &\mathbf{z} = \text{Upscale}(\mathbf{z}_{N_{i}}) + \text{Upscale}(\mathbf{z}_{N_{i-1}})\in\mathbb{R}^{M \times N_0 \times d}, \\
  \end{aligned}
\end{equation}
where $\mathbf{z}_{N_i}$ denotes the feature at scale $N_i$, and $\mathbf{hp}_{N_i}$ denotes the average-pooled head patch at scale $N_i$. Through this design, we encode multi-scale dynamics with shared weights, improving prediction accuracy while maintaining simplicity.

\section{Experiments}
\subsection{Dataset and Setup}
\textbf{Dataset:} We curate and benchmark Global Station Weather Forecasting (GSWF) tasks on \href{https://github.com/taohan10200/WEATHER-5K}{\textbf{Weather-5K}}, the largest up-to-date global station weather benchmark. It records hourly in-situ measurements of temperature, dew point, wind direction, wind rate, and sea-level pressure from 5k stations worldwide from 2014 to 2024. Although only Weather-5K is used in our experiments, it integrates and is quality-controlled against major station weather datasets, including HadISD, ICOADS, and others, making it sufficient to support reliable validation of model performance.

Beyond natural weather patterns, local weather systems are often highly coupled with human activities. We further validate our method on long-term explicit coupling patterns between weather systems and human water-regulation behaviors using \href{https://clp.engr.scu.edu/static/datasets/MCANN-datasets.zip}{\textbf{Hydrological Data}}. This dataset records hydrological observations from five reservoirs in the United States over 46 years. It contains both natural precipitation and anthropogenic regulation signals. Because extreme events are regulated and extreme patterns are often absent in the test years, only average accuracy is adopted for comparison.

Furthermore, to verify generalizability, we conduct evaluations on five \textbf{General Time Series Data} datasets that include human activities and energy dynamics. Since existing public datasets generally lack sufficiently long records with annual periodicity, we adopt a week-aligned historical structure with a historical range covering the past four weeks.

\textbf{Evaluation Protocol:} We comprehensively evaluate model performance in terms of average accuracy and extreme event capture using four metrics. For \textbf{accuracy}, we employ MAE/MSE to assess average deviations. For \textbf{extreme event prediction}, we use SEDI99.5th/SEDI90.0th to evaluate the prediction coverage of real-world extreme events. The computations are as follows:

\vspace{-10pt}
\begin{equation}
  \begin{aligned}
    \text{MAE} &= \frac{1}{C\times(T-T_o)} \sum \left(|\hat{\mathbf{Y}}_{H-1} - \mathbf{Y}_{H-1}| \right), \\
    \text{MSE} &= \frac{1}{C\times(T-T_o)} \sum \left((\hat{\mathbf{Y}}_{H-1} - \mathbf{Y}_{H-1})^2 \right), \\
  \end{aligned}
\end{equation}

\begin{equation}
  \begin{aligned}
    \text{SEDI} &= \frac{1}{\sum \mathbb{I}(\mathbf{Y}_{H-1}<\mathbf{q}^p_{\text{low}}) + \sum \mathbb{I}(\mathbf{Y}_{H-1}>\mathbf{q}^p_{\text{up}})} \Bigg( \\
      &\quad \sum \mathbb{I}(\hat{\mathbf{Y}}_{H-1}<\mathbf{q}^p_{\text{low}} \cap \mathbf{Y}_{H-1}<\mathbf{q}^p_{\text{low}}) + {} \\
    &\quad \sum \mathbb{I}(\hat{\mathbf{Y}}_{H-1}>\mathbf{q}^p_{\text{up}} \cap \mathbf{Y}_{H-1}>\mathbf{q}^p_{\text{up}}) \Bigg)\\
  \end{aligned}
\end{equation}
where $\mathbf{q}^p_{\text{low}}$ and $\mathbf{q}^p_{\text{up}}$ are lower and upper $p$th percentiles, respectively, and $p$ is set to $99.5/90.0$. $\mathbb{I}(\cdot)$ is indicator function.

\textbf{Baselines:} To comprehensively validate the superiority of our method, in addition to GSWF-specific models marked by $\star$, we collect state-of-the-art multivariate time series forecasting methods and retrain them on the GSWF task. Notably, cross-parameter-scale foundation model (Chronos2 120M), period-aware and memory-enhanced models which are relevant to our method are specifically included in this comparasion, marked by $\heartsuit$, $\lozenge$ and $\dag$, respectively. Moreover, we also introduce domain-specific methods, as marked by $\spadesuit$ and $\clubsuit$. For the missing-observation scenario, we compare with Merlin~\cite{yu2025merlin}, a recent noise-robust forecasting method.

\subsection{Overall performance}

\begin{table*}[!t]
  \centering
  \caption{Accuracy and Extreme Event Capture Comparison on \textbf{WEATHER DATA} (Year-over-year)\vspace{-5pt}}
  \label{tab:main_result_Weather-5K}
  \scriptsize
  \setlength{\tabcolsep}{1pt}
  \renewcommand{\arraystretch}{0.82}
  \begin{tabular*}{\textwidth}{@{\extracolsep{\fill}}@{}p{6.6em}@{}c@{}ccccccccc}
    \toprule
    \multirow{2}{*}{Baselines} & \multirow{2}{*}{Window} &
    \multicolumn{2}{c}{Temperature} & \multicolumn{2}{c}{Dewpoint} & \multicolumn{2}{c}{Wind Rate} & \multicolumn{1}{c}{Wind Direc.} & \multicolumn{2}{c}{Sea-Level Pressure} \\
    \cmidrule(lr){3-4} \cmidrule(lr){5-6} \cmidrule(lr){7-8} \cmidrule(lr){9-9} \cmidrule(lr){10-11}
    & & MAE/MSE$\downarrow$ & 99.5th/90.0th$\uparrow$ & MAE/MSE$\downarrow$ & 99.5th/90.0th$\uparrow$ & MAE/MSE$\downarrow$ & 99.5th/90.0th$\uparrow$ & MAE/MSE($\times 10^3$)$\downarrow$ & MAE/MSE$\downarrow$ & 99.5th/90.0th$\uparrow$\\
    \midrule
    \multirow{4}{*}{
      \begin{tabular}[c]{@{}l@{}}HRES~$\spadesuit$\\\cite{ecmwf2024hres}
    \end{tabular}}
    & 24h & 1.76 / 7.39 & \textcolor{green}{39} / \textcolor{green}{73} & \textcolor{green}{1.85} / \textcolor{green}{7.94} & \textcolor{green}{50.9} / 86.5 & \textcolor{green}{1.48} / \textcolor{green}{4.53} & \textcolor{green}{13.1} / \textcolor{green}{38.9} & \textcolor{green}{63.8} / 7.15 & 0.86 / 2.68 & 77.7 / 88.1 \\
    & 72h & 1.87 / 8.01 & \textcolor{green}{39.2} / \textcolor{green}{72.4} & 1.94 / 8.48 & 42.7 / 82.3 & 1.52 / 4.76 & \textcolor{green}{12} / \textcolor{green}{37.2} & 72.4 / 8.21 & 1.06 / 3.31 & 74 / 86.6 \\
    & 120h & 1.99 / 8.79 & \textcolor{green}{38.8} / \textcolor{green}{71.5} & 2.14 / 10.9 & 35.4 / 76.4 & \textcolor{green}{1.58} / \textcolor{green}{5.11} & 10.8 / 35.1 & \textcolor{green}{75.4} / \textcolor{green}{8.64} & 1.38 / 5.15 & 68.3 / 83.6 \\
    & 240h & 2.32 / 10.4 & \textcolor{green}{33.6} / \textcolor{green}{67} & 1.73 / 6.65 & 22.3 / 67.1 &  1.19 / 2.92 & \textcolor{green}{7.78} / 28.9 &\textcolor{green}{77.4} / \textcolor{green}{8.87} & 1.90 / 8.62 & 49.3 / 70.8 \\
    \cmidrule(lr){1-11}
    \multirow{4}{*}{
      \begin{tabular}[c]{@{}l@{}}FEDformer\\\cite{zhou2022fedformer}
    \end{tabular}} & 24h & 2.03 / 14.3 & 27.8 / 60.8 & 2.13 / 11.8 & 33.5 / 64.4 & 1.41 / 4.74 & 1.92 / 13.12 & 69.3 / 7.82 & 2.18 / 12.4 & 38.7 / 63.8 \\
    & 72h & 3.03 / 19.1 & 15.1 / 52.5 & 3.18 / 21.1 & 14.3 / 46 & 1.61 / 5.39 & 1.87 / 12.4 & 76.2 / 9.31 & 4.45 / 40.4 & 16.1 / 45.2 \\
    & 120h & 3.3 / 21.4 & 13.8 / 52.3 & 3.43 / 26.1 & 10.3 / 42.8 & 1.7 / 6.24 & 1.25 / 10.2 & 79 / 9.86 & 5.26 / 52.4 & 9.17 / 28.8 \\
    & 240h & 3.83 / 28.9 & 8.22 / 46.2 & 4.09 / 33.4 & 5.84 / 37.9 & 1.82 / 7.11 & 0.63 / 8.12 & 83.8 / 11.2 & 6.39 / 82.8 & 5.8 / 27.2 \\
    \cmidrule(lr){1-11}
    \multirow{4}{*}{
      \begin{tabular}[c]{@{}l@{}}PatchTST\\\cite{nie2023time}
    \end{tabular}}
    & 24h & 2.14 / 10.5 & 17.2 / 55.9 & 2.21 / 12.1 & 27.4 / 63.3 & 1.47 / 4.62 & 1.94 / 16.3 & 67.7 / 8.36 & 3 / 20.5 & 33.6 / 60.6 \\
    & 72h & 3.11 / 19.8 & 8.26 / 46.7 & 3.11 / 20.6 & 14.1 / 49.8 & 1.66 / 5.72 & 0.86 / 11.6 & 76.1 / 9.66 & 4.93 / 52.5 & 14.4 / 39.7 \\
    & 120h & 3.19 / 20.9 & 7.71 / 43.8 & 3.48 / 25.5 & 11.1 / 44.9 & 1.72 / 6.33 & 0.74 / 9.6 & 79.2 / 10.2 & 5.64 / 66.7 & 7.63 / 32.5 \\
    & 240h & 3.99 / 30 & 4.48 / 38.9 & 4.01 / 32 & 7.52 / 39.3 & 1.76 / 6.7 & 0.33 / 8.82 & 81.4 / 10.5 & 6.47 / 84.2 & 3.46 / 24.5 \\
    \cmidrule(lr){1-11}
    \multirow{4}{*}{
      \begin{tabular}[c]{@{}l@{}}iTransformer\\\cite{liu2023itransformer}
    \end{tabular}}
    & 24h & 1.93 / 9.34 & 28 / 64.4 & 2.03 / 9.15 & 29.8 / 65.1 & 1.44 / 4.53 & 3.54 / 19 & 67.2 / 8.35 & 2.21 / 12.3 & 48.2 / 67.7 \\
    & 72h & 2.79 / 16.3 & 20.9 / 56.9 & 2.96 / 19.4 & 16.3 / 50.8 & 1.63 / 5.34 & 1.51 / 12.6 & 75 / 9.22 & 4.33 / 42.9 & 23.2 / 47.2 \\
    & 120h & 3.09 / 21.2 & 14.6 / 52.1 & 3.34 / 24.9 & 9.78 / 44.7 & 1.7 / 6.17 & 1.42 / 10.3 & 77.8 / 9.64 & 5.18 / 58.5 & \textcolor{red}{13} / 36.6 \\
    & 240h & 3.73 / 27.5 & 9.54 / 46.3 & 3.98 / 31.8 & 7.23 / 39.4 & 1.75 / 6.49 & 0.75 / 9.11 & 80.8 / 10.2 & 6.06 / 75.7 & 5.67 / 26.4 \\
    \cmidrule(lr){1-11}
    \multirow{4}{*}{
      \begin{tabular}[c]{@{}l@{}}Pyraformer\\\cite{liu2022pyraformer}
    \end{tabular}}
    & 24h & \textcolor{blue}{1.85} / \textcolor{blue}{7.19} & 22.6 / 64.8 & 2.02 / 8.17 & 21 / 61.5 & 1.42 / 4.28 & 2.42 / 13.6 & 63.8 / \textcolor{blue}{7.15} & 2.02 / \textcolor{blue}{10.9} & 32.1 / 62.7 \\
    & 72h & \textcolor{blue}{2.58} / 14.7 & 15.9 / 58.3 & \textcolor{blue}{2.85} / \textcolor{blue}{17.3} & 11.4 / 46.9 & \textcolor{blue}{1.58} / \textcolor{red}{4.87} & 0.75 / 8.34 & \textcolor{red}{74.1} / \textcolor{red}{8.34} & \textcolor{blue}{3.87} / \textcolor{blue}{35.5} & 14.5 / 34.4 \\
    & 120h & 2.99 / 18.2 & 9.5 / 52.6 & 3.12 / 20.4 & 5.21 / 39.1 & 1.62 / \textcolor{blue}{5.51} & 0.49 / 6.33 & 77 / \textcolor{blue}{8.73} & \textcolor{blue}{4.59} / \textcolor{blue}{47.2} & 7.67 / 22.9 \\
    & 240h & 3.49 / 23.8 & 6.56 / 47.2 & 3.72 / 27.1 & 3.37 / 31.6 & 1.66 / 5.78 & 0.31 / 5.19 & 79.3 / 9.09 & 5.38 / 59 & 3.08 / 12.6 \\
    \cmidrule(lr){1-11}
    \multirow{4}{*}{
      \begin{tabular}[c]{@{}l@{}}DLinear\\\cite{zeng2023transformers}
    \end{tabular}}
    & 24h & 2.79 / 14.7 & 8.66 / 42.8 & 2.54 / 13.3 & 11.1 / 44.5 & 1.46 / 4.52 & 1.51 / 14.3 & 67.4 / 8.33 & 3.16 / 22.1 & 14.4 / 46.6 \\
    & 72h & 3.65 / 24.2 & 5.06 / 28.1 & 3.58 / 24.1 & 5.22 / 28 & 1.6 / 5.86 & 0.65 / 7.9 & 75 / \textcolor{blue}{9.17} & 4.61 / 49.3 & 5.21 / 27.2 \\
    & 120h & 3.99 / 29 & 2.76 / 23.2 & 3.99 / 29.4 & 3.26 / 21.9 & 1.69 / 5.92 & 0.43 / 5.68 & 77.4 / 9.46 & 5.26 / 57.1 & 2.93 / 20.1 \\
    & 240h & 4.43 / 35 & 1.87 / 18 & 4.45 / 35.4 & 2.45 / 16.2 & 1.72 / 6.18 & 0.27 / 4.54 & 79.8 / 9.79 & 5.86 / 68.9 & 1.2 / 12.8 \\
    \cmidrule(lr){1-11}
    \multirow{4}{*}{
      \begin{tabular}[c]{@{}l@{}}Mamba\\\cite{gu2023mamba}
    \end{tabular}}
    & 24h & 1.93 / 9.81 & 30.1 / 65.7 & 2.11 / 11.9 & 26.5 / 63.8 & 1.5 / 4.77 & 1.58 / 13.8 & 70.9 / 8.32 & 2.23 / 13.2 & 37 / 62.8 \\
    & 72h & 2.9 / 17.1 & 17.1 / 56 & 2.99 / 19.3 & 15.6 / 51.1 & 1.68 / 5.34 & 1.61 / 10.4 & 75.4 / 10.83 & 4.37 / 43 & 17.2 / 43.1 \\
    & 120h & 3.11 / 19.4 & 9.77 / 50.7 & 3.24 / 22.9 & 8.56 / 43.5 & 1.61 / \textcolor{blue}{5.51} & 0.77 / 7.53 & 77.7 / 9.05 & 5.12 / 55.3 & 8.69 / 30 \\
    & 240h & 3.68 / 30.2 & 4.26 / 43.7 & 3.94 / 32.2 & 2.97 / 32 & 1.71 / 5.97 & 0.5 / 5.27 & \textcolor{blue}{77.3} / 9.58 & 5.57 / \textcolor{blue}{53.9} & 3.48 / 16.8 \\
    \cmidrule(lr){1-11}
    \multirow{4}{*}{
      \begin{tabular}[c]{@{}l@{}}Chronos2~$\heartsuit$\\(120M)~\cite{ansari2025chronos2}
    \end{tabular}}
    & 24h & 2.05 / 9.5 & \textcolor{blue}{36.3} / \textcolor{blue}{66.3} & 2.06 / 12.3 & 31.3 / 65.2 & 1.47 / 4.51 & 8.81 / 24.1 & 64.4 / 8.65 & 2.13 / 11.3 & 52.4 / \textcolor{blue}{71.8} \\
    & 72h & 2.83 / 17.5 & 23.3 / 57.5 & 2.95 / 22.3 & 18.7 / 50.3 & \textcolor{red}{1.55} / \textcolor{blue}{5.02} & 5.24 / 16.1 & 75.9 / 10.9 & 4.17 / 39.5 & 24.1 / 46.3 \\
    & 120h & 3.32 / 23.3 & 16.8 / 27.2 & 3.52 / 29.5 & 10.3 / 19.6 & 1.64 / 5.53 & 4.48 / 6.03 & 78.4 / 10.1 & 4.95 / 54.9 & 11.9 / 17.6 \\
    & 240h & 3.63 / 26.5 & 12.8 / 48.7 & 3.89 / 31.1 & \textcolor{blue}{9.52} / 38.9 & 1.73 / 6.65 & 4.02 / 12.78 & 81.8 / 11.4 & 5.55 / 64.3 & 7.29 / 27.6 \\
    \cmidrule(lr){1-11}
    \multirow{4}{*}{
      \begin{tabular}[c]{@{}l@{}}WSSM~$\star$\\\cite{yang2025wssm}
    \end{tabular}}
    & 24h & \textcolor{red}{1.75} / \textcolor{red}{7.16} & 33.8 / 66.2 & \textcolor{blue}{1.99} / \textcolor{blue}{7.91} & 28.3 / 64.4 & \textcolor{blue}{1.39} / \textcolor{blue}{4.13} & 8.44 / 20.9 & \textcolor{blue}{62.9} / \textcolor{red}{7.06} & \textcolor{blue}{2.01} / \textcolor{blue}{10.9} & 37.1 / 59.6 \\
    & 72h & 2.65 / 15.5 & 22.1 / \textcolor{blue}{62.4} & 2.89 / 18.1 & 21.1 / 52.2 & 1.68 / 5.78 & \textcolor{blue}{8.26} / \textcolor{blue}{20} & 75.9 / 10.8 & 3.91 / 36.4 & 28.2 / \textcolor{blue}{55.7} \\
    & 120h & \textcolor{blue}{2.97} / \textcolor{blue}{17.8} & 18.22 / \textcolor{blue}{58.2} & \textcolor{blue}{3.08} / \textcolor{blue}{19.7} & \textcolor{blue}{13.57} / 45.2 & \textcolor{blue}{1.59} / 5.91 & \textcolor{blue}{4.98} / 12.4 & \textcolor{blue}{75.8} / 8.78 & 4.83 / 54.0 & 10.5 / 37.9 \\
    & 240h & \textcolor{blue}{3.44} / \textcolor{blue}{23} & \textcolor{blue}{14.7} / \textcolor{blue}{55.3} & 3.6 / \textcolor{blue}{25.9} & \textcolor{red}{12.9} / 42.1 & \textcolor{blue}{1.62} / \textcolor{blue}{5.57} & \textcolor{blue}{4.89} / \textcolor{blue}{18.1} & 78.9 / \textcolor{blue}{9.08} & 5.57 / 63.9 & \textcolor{red}{8.91} / \textcolor{blue}{31.2} \\
    \cmidrule(lr){1-11}
    \multirow{4}{*}{
      \begin{tabular}[c]{@{}l@{}}Corrformer~$\star$\\\cite{wu2023corrformer}
    \end{tabular}}
    & 24h & 2.03 / 9.72 & 34.6 / 62.3 & 2.14 / 10.1 & \textcolor{blue}{35} / \textcolor{blue}{67.7} & 1.49 / 4.56 & \textcolor{blue}{10.9} / \textcolor{blue}{26.7} & 67.7 / 8.9 & 2.27 / 17.1 & \textcolor{blue}{52.6} / 70.7 \\
    & 72h & 2.86 / 17.8 & \textcolor{blue}{28.1} / 59.7 & 3 / 19.8 & \textcolor{blue}{23.2} / \textcolor{blue}{58.3} & 1.63 / 5.67 & 5.97 / 18.5 & 76.4 / 10.9 & 4.24 / 40 & \textcolor{blue}{29.2} / 53.1 \\
    & 120h & 3.12 / 19.7 & \textcolor{blue}{20.9} / 55.9 & 3.36 / 23.9 & 11.2 / \textcolor{blue}{50.9} & 1.67 / 5.55 & 2.6 / 11.5 & 76.6 / 9.74 & 4.9 / 58.9 & \textcolor{blue}{12.4} / \textcolor{blue}{40} \\
    & 240h & 3.51 / \textcolor{blue}{23} & 13.3 / 53.8 & \textcolor{blue}{3.58} / 27.2 & 8.05 / \textcolor{blue}{43.3} & 1.66 / 6.01 & 4.85 / 17.9 & 78.6 / 9.66 & \textcolor{blue}{5.37} / 61.1 & 6.15 / 30.4 \\
    \cmidrule(lr){1-11}
    \multirow{4}{*}{
      \begin{tabular}[c]{@{}l@{}}Autoformer~$\lozenge$\\\cite{wu2021autoformer}
    \end{tabular}}
    & 24h & 2.06 / 10.3 & 27.7 / 56.3 & 2.4 / 13.2 & 31.8 / 66.3 & 1.42 / 4.49 & 9.25 / 23.6 & 67.2 / 7.88 & 2.43 / 14.78 & 41.5 / 64.7 \\
    & 72h & 2.87 / 17.6 & 16.8 / 50.4 & 3.08 / 20.4 & 21.4 / 57.2 & 1.66 / 5.42 & 5.75 / 17.2 & 75.5 / 9.27 & 4.52 / 47 & 22.3 / 47.8 \\
    & 120h & 3.85 / 21.9 & 11.6 / 42.9 & 3.8 / 29.9 & 11.3 / 48.6 & 1.73 / 5.9 & 3.94 / \textcolor{blue}{14.2} & 78.6 / 10.9 & 6.02 / 74.2 & 11.6 / 35.3 \\
    & 240h & 4.29 / 32.5 & 1.68 / 25.4 & 3.88 / 30 & 4.19 / 36.8 & 1.77 / 6.33 & 0.37 / 6.71 & 79.6 / 9.88 & 6.66 / 63.5 & 2.25 / 22.4 \\
    \cmidrule(lr){1-11}
    \multirow{4}{*}{
      \begin{tabular}[c]{@{}l@{}}PhaseFormer~$\lozenge$\\\cite{niu2026phaseformer}
    \end{tabular}}
    & 24h & 2.19 / 12.1 & 9.86 / 48.6 & 2.74 / 16.1 & 17.2 / 57.9 & 1.53 / 4.75 & 0.944 / 12.2 & 69.5 / 7.98 & 3.16 / 25.3 & 20.8 / 50 \\
    & 72h & 2.65 / 19.6 & 5.84 / 40.3 & 3.41 / 23.8 & 10.9 / 47.8 & 1.7 / 6.06 & 1.01 / 9.9 & 78.3 / 10.1 & 4.78 / 51.4 & 9.08 / 35.4 \\
    & 120h & 3.89 / 22.2 & 4.21 / 37.7 & 3.82 / 29.6 & 9.31 / 44.5 & 1.79 / 6.65 & 0.972 / 8.24 & 80.9 / 10.5 & 6.14 / 75.9 & 5.02 / 28.9 \\
    & 240h & 4.21 / 32.8 & 3.2 / 33.2 & 4.16 / 34.6 & 5.81 / 39 & 1.81 / 7.02 & 0.488 / 7.57 & 82.6 / 10.8 & 6.63 / 88.3 & 2.22 / 23.3 \\
    \cmidrule(lr){1-11}
    \multirow{4}{*}{
      \begin{tabular}[c]{@{}l@{}}SparseTSF~$\lozenge$\\\cite{lin2024sparsetsf}
    \end{tabular}}
    & 24h & 2.52 / 14.8 & 7.87 / 42 & 2.9 / 17.5 & 16.7 / 55.9 & 1.68 / 5.75 & 1.6 / 11.2 & 75.5 / 9.59 & 4.4 / 38.3 & 17.7 / 45 \\
    & 72h & 2.96 / 25.8 & 5.3 / 34.8 & 3.53 / 25.1 & 10.3 / 46.1 & 1.78 / 6.41 & 0.709 / 9.07 & 80.2 / 10.4 & 5.67 / 64.7 & 8.05 / 32.9 \\
    & 120h & 4.15 / 31.2 & 4.8 / 34 & 3.8 / 29.2 & 9.1 / 43.2 & 1.82 / 6.93 & 0.648 / 7.84 & 82.1 / 10.7 & 6.16 / 75.8 & 4.76 / 28.6 \\
    & 240h & 4.49 / 36 & 3.2 / 30.1 & 4.17 / 34.1 & 6.27 / 38.5 & 1.85 / 7.23 & 0.346 / 7.6 & 83.9 / 11 & 6.79 / 90.8 & 2.12 / 22.8 \\
    \cmidrule(lr){1-11}
    \multirow{4}{*}{
      \begin{tabular}[c]{@{}l@{}}CycleNet~\dag\\\cite{cyclenet}
    \end{tabular}}
    & 24h & 2.06 / 11.2 & 11.7 / 49.3 & 2.65 / 15 & 21.4 / 59.2 & 1.61 / 5.36 & 4.71 / 16.3 & 68.3 / 7.67 & 2.85 / 20.7 & 21.5 / 50.2 \\
    & 72h & \textcolor{red}{2.54} / \textcolor{red}{14.2} & 7.11 / 43.3 & 3.37 / 23.3 & 12.7 / 48.4 & 1.74 / 6.23 & 3.37 / 12.8 & 80.2 / 10.7 & 4.93 / 51.4 & 10.1 / 35.4 \\
    & 120h & 3.78 / 20.3 & 6.4 / 41 & 3.69 / 27.9 & 10.1 / 44.4 & 1.79 / 6.8 & 3.37 / 11.5 & 82.7 / 11.1 & 5.89 / 70.8 & 5.67 / 30.3 \\
    & 240h & 4.13 / 31.6 & 3.86 / 36.9 & 4.08 / 32.9 & 7.13 / 39.5 & 1.82 / 7.12 & 1.96 / 10.4 & 84.5 / 11.4 & 6.6 / 86.7 & 2.75 / 23.6 \\
    \cmidrule(lr){1-11}
    \multirow{4}{*}{
      \begin{tabular}[c]{@{}l@{}}TQNet~\dag\\\cite{lin2025tqnet}
    \end{tabular}}
    & 24h & 1.94 / 12.6 & 17.5 / 58.3 & 2.24 / 11.8 & 28.9 / 65.1 & 1.47 / 4.62 & 1.87 / 16.3 & 67.2 / 7.33 & 2.82 / 18.6 & 40.1 / 64.1 \\
    & 72h & 2.59 / 19.6 & 8.84 / 49.5 & 3.11 / 20.7 & 15.1 / 51.6 & 1.62 / 5.56 & 1.56 / 11.8 & 75.9 / 9.63 & 4.7 / 48.5 & 15.6 / 41.9 \\
    & 120h & 3.55 / 24.6 & 7.03 / 46.5 & 3.59 / 27 & 11.4 / 47.2 & 1.71 / 6.2 & 1.03 / 9.59 & 79.2 / 10.2 & 5.54 / 65 & 8.79 / 32.8 \\
    & 240h & 3.9 / 29.2 & 4.54 / 41.1 & 4 / 32.6 & 7.44 / 40.2 & 1.74 / 6.65 & 0.633 / 8.82 & 81.2 / 10.5 & 6.23 / 80.2 & 3.68 / 24.8 \\
    \cmidrule(lr){1-11}
    \multirow{4}{*}{\textbf{TSSM (ours)}}
    & 24h & 2.36 / 11.3 & \textcolor{red}{83.7} / \textcolor{red}{87.6} & \textcolor{red}{1.62} / \textcolor{red}{7.01} & \textcolor{red}{53.6} / \textcolor{red}{77.5} & \textcolor{red}{1.31} / \textcolor{red}{3.91} & \textcolor{red}{33.4} / \textcolor{red}{49.7} & \textcolor{red}{60.3} / 7.96 & \textcolor{red}{1.77} / \textcolor{red}{8.54} & \textcolor{red}{67.5} / \textcolor{red}{82.2} \\
    & 72h & 2.61 / \textcolor{blue}{14.3} & \textcolor{red}{63.3} / \textcolor{red}{81.6} & \textcolor{red}{2.46} / \textcolor{red}{15.2} & \textcolor{red}{31.2} / \textcolor{red}{63.7} & 1.61 / 5.72 & \textcolor{red}{30.2} / \textcolor{red}{39.3} & \textcolor{blue}{74.5} / 10.6 & \textcolor{red}{3.11} / \textcolor{red}{25} & \textcolor{red}{39.9} / \textcolor{red}{61.1} \\
    & 120h & \textcolor{red}{2.54} / \textcolor{red}{13.7} & \textcolor{red}{45.6} / \textcolor{red}{73.3} & \textcolor{red}{2.51} / \textcolor{red}{13.2} & \textcolor{red}{16.4} / \textcolor{red}{57.5} & \textcolor{red}{1.44} / \textcolor{red}{4.46} & \textcolor{red}{9.58} / \textcolor{red}{22.5} & \textcolor{red}{68.7} / \textcolor{red}{8.04} & \textcolor{red}{3.72} / \textcolor{red}{28.6} & 12.3 / \textcolor{red}{44.8} \\
    & 240h & \textcolor{red}{3.04} / \textcolor{red}{17.2} & \textcolor{red}{47} / \textcolor{red}{74.5} & \textcolor{red}{2.83} / \textcolor{red}{16.2} & 6.4 / \textcolor{red}{45.6} & \textcolor{red}{1.49} / \textcolor{red}{4.4} & \textcolor{red}{13.9} / \textcolor{red}{24.6} & \textcolor{red}{70.5} / \textcolor{red}{8.43} & \textcolor{red}{3.69} / \textcolor{red}{28.1} & \textcolor{blue}{8.26} / \textcolor{red}{36.5} \\
    \bottomrule
  \end{tabular*}
  \\[2pt]
  \makebox[\textwidth][l]{\small Note: $\spadesuit$: Physics-based method; $\heartsuit$: Foundation Model $\star$: GSWF AI-methods; $\lozenge$: Period-aware AI-methods; \dag: Memory-enhanced AI-methods.}
  \\[1pt]
  \makebox[\textwidth][l]{\small Best and second-best results are highlighted in \textcolor{red}{red} and \textcolor{blue}{blue}, respectively.}
  \\[1pt]
  \makebox[\textwidth][l]{\small Metrics where our method outperforms operational ECMWF-HRES are marked in \textcolor{green}{green}.}
  \vspace{-15pt}
\end{table*}

\begin{table*}[!t]
  \centering
  \caption{Accuracy Comparison on \textbf{HYDROLOGY DATA} (Year-over-year) \vspace{-5pt}}
  \label{tab:main_result_hydrology}
  \scriptsize
  \setlength{\tabcolsep}{2pt}
  \renewcommand{\arraystretch}{0.82}
  \begin{tabular*}{\textwidth}{@{\extracolsep{\fill}}@{}p{5.2em}@{}c@{}cccccccccc}
    \toprule
    \multirow{2}{*}{Baselines} & \multirow{2}{*}{Window} &
    \multicolumn{2}{c}{Almaden} & \multicolumn{2}{c}{Coyote} & \multicolumn{2}{c}{Lexington} & \multicolumn{2}{c}{Stevens Creek} & \multicolumn{2}{c}{Vasona} \\
    \cmidrule(lr){3-4} \cmidrule(lr){5-6} \cmidrule(lr){7-8} \cmidrule(lr){9-10} \cmidrule(lr){11-12}
    & & MAE$\downarrow$ & MSE$\downarrow$ & MAE$\downarrow$ & MSE$\downarrow$ & MAE$\downarrow$ & MSE$\downarrow$ & MAE$\downarrow$ & MSE$\downarrow$ & MAE$\downarrow$ & MSE$\downarrow$\\
    \midrule
    \multirow{4}{*}{PatchTST} & 24h & 1.3e-2 & 6.1e-3 & \textcolor{blue}{5.0e-2} & \textcolor{blue}{2.0e-3} & 1.6e-2 & 3.3e-3 & \textcolor{blue}{9.0e-3} & 1.1e-3 & 8.7e-3 & 4.8e-4 \\
    & 72h & 3.1e-2 & 1.2e-2 & 9.5e-2 & \textcolor{blue}{4.0e-2} & 4.4e-2 & 1.5e-2 & 3.0e-2 & 1.7e-2 & \textcolor{blue}{2.5e-2} & 3.8e-3 \\
    & 120h & \textcolor{blue}{2.9e-2} & \textcolor{red}{4.3e-3} & 2.6e-1 & 8.2e-2 & 5.9e-2 & 1.7e-2 & \textcolor{red}{2.0e-2} & \textcolor{blue}{1.8e-3} & \textcolor{blue}{2.8e-2} & 5.7e-3 \\
    & 240h & \textcolor{blue}{1.5e-1} & \textcolor{blue}{4.1e-2} & 3.3e-1 & 1.3e-1 & \textcolor{blue}{2.8e-1} & \textcolor{blue}{1.4e-1} & \textcolor{blue}{8.3e-2} & \textcolor{blue}{1.2e-2} & 1.9e-1 & 6.2e-2 \\
    \cmidrule(lr){1-12}
    \multirow{4}{*}{iTransformer} & 24h & \textcolor{blue}{1.2e-2} & 5.9e-3 & 5.6e-2 & 2.5e-2 & 1.7e-2 & 3.7e-3 & 1.0e-2 & 1.6e-3 & \textcolor{blue}{8.2e-3} & \textcolor{red}{3.8e-4} \\
    & 72h & 2.7e-2 & 1.0e-2 & 1.0e-1 & 4.5e-2 & 4.6e-2 & 1.7e-2 & 3.1e-2 & 2.2e-2 & \textcolor{red}{2.4e-2} & \textcolor{red}{3.5e-3} \\
    & 120h & \textcolor{blue}{2.9e-2} & \textcolor{red}{4.3e-3} & 2.8e-1 & 9.4e-2 & 5.9e-2 & 1.7e-2 & \textcolor{red}{2.0e-2} & \textcolor{blue}{1.8e-3} & 3.0e-2 & 5.8e-3 \\
    & 240h & \textcolor{blue}{1.5e-1} & \textcolor{blue}{4.1e-2} & 3.5e-1 & 1.4e-1 & \textcolor{blue}{2.8e-1} & \textcolor{blue}{1.4e-1} & \textcolor{blue}{8.3e-2} & \textcolor{red}{1.1e-2} & 1.9e-1 & 6.2e-2 \\
    \cmidrule(lr){1-12}
    \multirow{4}{*}{
      \begin{tabular}[c]{@{}l@{}}MCANN~$\clubsuit$\\\cite{li2025mcann}
    \end{tabular}} & 24h & 1.3e-2 & \textcolor{red}{6.0e-4} & 5.1e-2 & 2.4e-2 & \textcolor{blue}{1.5e-2} & \textcolor{blue}{3.1e-3} & \textcolor{blue}{9.0e-3} & \textcolor{blue}{8.2e-4} & 9.4e-3 & 5.9e-4 \\
    & 72h & 2.9e-2 & 1.0e-2 & \textcolor{blue}{9.4e-2} & \textcolor{blue}{4.0e-2} & \textcolor{blue}{4.3e-2} & \textcolor{blue}{1.4e-2} & 3.2e-2 & 2.0e-2 & \textcolor{red}{2.4e-2} & \textcolor{blue}{3.7e-3} \\
    & 120h & \textcolor{red}{2.8e-2} & \textcolor{red}{4.3e-3} & 2.0e-1 & 5.1e-2 & 5.8e-2 & 1.8e-2 & \textcolor{red}{2.0e-2} & 1.9e-3 & \textcolor{blue}{2.8e-2} & \textcolor{blue}{5.3e-3} \\
    & 240h & \textcolor{blue}{1.5e-1} & \textcolor{blue}{4.1e-2} & 2.8e-1 & 9.4e-2 & \textcolor{blue}{2.8e-1} & \textcolor{blue}{1.4e-1} & \textcolor{blue}{8.3e-2} & \textcolor{red}{1.1e-2} & 1.9e-1 & 6.2e-2 \\
    \cmidrule(lr){1-12}
    \multirow{4}{*}{SparseTSF $\lozenge$} & 24h & 3.1e-2 & 1.5e-2 & 8.7e-2 & 4.2e-2 & 3.0e-2 & 6.0e-3 & 2.1e-2 & 2.8e-3 & 2.3e-2 & 3.4e-3 \\
    & 72h & 5.0e-2 & 2.0e-2 & 1.2e-1 & 6.1e-2 & 5.9e-2 & 1.7e-2 & 4.4e-2 & 2.1e-2 & 4.2e-2 & 8.5e-3 \\
    & 120h & \textcolor{red}{2.8e-2} & \textcolor{red}{4.3e-3} & 1.9e-1 & 6.2e-2 & 5.8e-2 & 1.7e-2 & \textcolor{red}{2.0e-2} & \textcolor{blue}{1.8e-3} & 2.9e-2 & 5.4e-3 \\
    & 240h & \textcolor{blue}{1.5e-1} & \textcolor{blue}{4.1e-2} & \textcolor{blue}{1.4e-1} & \textcolor{blue}{4.0e-2} & \textcolor{blue}{2.8e-1} & \textcolor{blue}{1.4e-1} & \textcolor{blue}{8.3e-2} & \textcolor{blue}{1.2e-2} & \textcolor{blue}{1.8e-1} & \textcolor{blue}{6.0e-2} \\
    \cmidrule(lr){1-12}
    \multirow{4}{*}{TQNet \dag} & 24h & 1.4e-2 & 6.3e-3 & 5.9e-2 & 2.9e-2 & 1.8e-2 & 3.9e-3 & 1.0e-2 & 1.2e-3 & 9.9e-3 & 6.0e-4 \\
    & 72h & \textcolor{blue}{2.6e-2} & \textcolor{red}{4.6e-3} & 1.4e-1 & 9.8e-2 & 6.0e-2 & 2.4e-2 & \textcolor{blue}{2.0e-2} & \textcolor{blue}{2.6e-3} & 2.8e-2 & 8.2e-3 \\
    & 120h & \textcolor{red}{2.8e-2} & 1.0e-2 & \textcolor{blue}{1.0e-1} & \textcolor{blue}{4.7e-2} & \textcolor{blue}{4.7e-2} & \textcolor{blue}{1.6e-2} & 3.1e-2 & 1.8e-2 & \textcolor{red}{2.6e-2} & \textcolor{red}{4.0e-3} \\
    & 240h & \textcolor{blue}{1.5e-1} & \textcolor{blue}{4.1e-2} & 4.1e-1 & 1.9e-1 & \textcolor{blue}{2.8e-1} & \textcolor{blue}{1.4e-1} & \textcolor{blue}{8.3e-2} & \textcolor{blue}{1.2e-2} & \textcolor{blue}{1.8e-1} & 6.1e-2 \\
    \cmidrule(lr){1-12}
    \multirow{4}{*}{\textbf{TSSM (ours)}} & 24h & \textcolor{red}{9.2e-3} & \textcolor{blue}{1.4e-3} & \textcolor{red}{2e-3} & \textcolor{red}{1.7e-5} & \textcolor{red}{3.5e-3} & \textcolor{red}{2.2e-4} & \textcolor{red}{7.7e-3} & \textcolor{red}{7.8e-4} & \textcolor{red}{2.5e-3} & \textcolor{blue}{4.3e-4} \\
    & 72h & \textcolor{red}{2.3e-2} & \textcolor{blue}{5.4e-3} & \textcolor{red}{1.6e-2} & \textcolor{red}{8.8e-4} & \textcolor{red}{1.2e-2} & \textcolor{red}{1.4e-4} & \textcolor{red}{1.2e-2} & \textcolor{red}{2.6e-4} & 3.2e-2 & 6.1e-3 \\
    & 120h & \textcolor{blue}{2.9e-2} & \textcolor{blue}{6.7e-3} & \textcolor{red}{1.8e-2} & \textcolor{red}{5.7e-4} & \textcolor{red}{2.4e-2} & \textcolor{red}{7.9e-4} & \textcolor{blue}{2.9e-2} & \textcolor{red}{1.7e-3} & 3.1e-2 & 6.2e-3 \\
    & 240h & \textcolor{red}{5.5e-2} & \textcolor{red}{1.8e-2} & \textcolor{red}{4e-2} & \textcolor{red}{1.1e-2} & \textcolor{red}{5.1e-2} & \textcolor{red}{1.7e-2} & \textcolor{red}{4.8e-2} & 1.4e-2 & \textcolor{red}{1e-1} & \textcolor{red}{3.7e-2} \\
    \bottomrule
  \end{tabular*}
  \\[2pt]
  \makebox[\textwidth][l]{\small Note: $\clubsuit$: Hydrology AI-methods $\lozenge$: period-aware AI-methods; \dag: memory-enhanced AI-methods.}
  \vspace{-10pt}
\end{table*}

\begin{table*}[!t]
  \centering
  \caption{Accuracy and Extreme Event Capture Comparison on \textbf{GENERAL TIME SERIES DATA} (Week-over-week) \vspace{-5pt}}
  \label{tab:main_result_TS}
  \scriptsize
  \setlength{\tabcolsep}{2pt}
  \renewcommand{\arraystretch}{0.82}
  \begin{tabular*}{\textwidth}{@{\extracolsep{\fill}}@{}p{5.2em}@{}c@{}cccccccccc}
    \toprule
    \multirow{2}{*}{Baselines} & \multirow{2}{*}{Window} &
    \multicolumn{2}{c}{ETTh1} & \multicolumn{2}{c}{ETTh2} & \multicolumn{2}{c}{Electricity} & \multicolumn{2}{c}{Traffic} & \multicolumn{2}{c}{Solar-Energy} \\
    \cmidrule(lr){3-4} \cmidrule(lr){5-6} \cmidrule(lr){7-8} \cmidrule(lr){9-10} \cmidrule(lr){11-12}
    & & MAE/MSE$\downarrow$ & 99.5th/90.0th$\uparrow$ & MAE/MSE$\downarrow$ & 99.5th/90.0th$\uparrow$ & MAE/MSE$\downarrow$ & 99.5th/90.0th$\uparrow$ & MAE/MSE$\downarrow$ & 99.5th/90.0th$\uparrow$ & MAE/MSE$\downarrow$ & 99.5th/90.0th$\uparrow$\\
    \midrule
    \multirow{4}{*}{PatchTST} & 24h & \textcolor{blue}{2.1e-1} / 8.2e-2 & \textcolor{blue}{29.7} / \textcolor{blue}{64.5} & \textcolor{blue}{2e-1} / \textcolor{blue}{8.5e-2} & 46.8 / \textcolor{blue}{66.2} & \textcolor{blue}{1.5e-1} / \textcolor{blue}{7e-1} & 8.71 / 22.7 & \textcolor{blue}{3.0e-1} / 3.6e-1 & 5.97 / 27.53 & \textcolor{blue}{3.1e-1} / \textcolor{blue}{3.3e-1} & 16.79 / 44.40 \\
    & 72h & \textcolor{blue}{3.3e-1} / \textcolor{blue}{2.1e-1} & 16.8 / 48.8 & \textcolor{blue}{3e-1} / \textcolor{blue}{1.7e-1} & \textcolor{blue}{18.7} / 50.1 & \textcolor{blue}{1.7e-1} / \textcolor{blue}{9.2e-1} & 6 / 19.8 & \textcolor{blue}{3.6e-1} / 4.7e-1 & 3.12 / 17.46 & \textcolor{blue}{5.4e-1} / 8.9e-1 & 3.27 / 12.77 \\
    & 120h & \textcolor{blue}{3.1e-1} / \textcolor{blue}{1.8e-1} & 8.21 / 44.4 & \textcolor{blue}{3.4e-1} / 2.1e-1 & \textcolor{blue}{35.1} / \textcolor{blue}{43.2} & \textcolor{blue}{1.7e-1} / \textcolor{blue}{1.04} & 6.49 / 20.3 & \textcolor{blue}{3.5e-1} / \textcolor{blue}{4.7e-1} & 3.38 / 18.74 & \textcolor{blue}{4.5e-1} / 6.8e-1 & 3.29 / 16.18 \\
    & 240h & \textcolor{blue}{8e-1} / \textcolor{blue}{8.4e-1} & \textcolor{blue}{0} / \textcolor{blue}{35.2} & 6.8e-1 / 6.4e-1 & \textcolor{blue}{0} / \textcolor{blue}{0} & \textcolor{blue}{1.6e-1} / \textcolor{blue}{8e-1} & 5.28 / 18.7 & \textcolor{red}{3.4e-1} / \textcolor{blue}{4.4e-1} & 4.51 / 22.96 & \textcolor{blue}{4.7e-1} / 7.6e-1 & 3.41 / 14.32 \\
    \cmidrule(lr){1-12}
    \multirow{4}{*}{iTransformer} & 24h & 2.2e-1 / \textcolor{red}{9.4e-3} & 29.3 / 63.2 & 2.1e-1 / \textcolor{blue}{8.5e-2} & \textcolor{blue}{53.2} / \textcolor{blue}{66.2} & \textcolor{blue}{1.5e-1} / 7.1e-1 & 15.5 / 30.4 & \textcolor{red}{2.9e-1} / \textcolor{blue}{3.5e-1} & \textcolor{blue}{15.53} / \textcolor{blue}{44.17} & 3.4e-1 / 3.8e-1 & \textcolor{blue}{39.81} / \textcolor{blue}{51.44} \\
    & 72h & 3.4e-1 / 2.2e-1 & 16.7 / 41.3 & 3.1e-1 / 1.8e-1 & 16.4 / 46.4 & 1.8e-1 / 1.01 & 12 / 26.8 & \textcolor{blue}{3.6e-1} / \textcolor{blue}{4.6e-1} & \textcolor{blue}{9.12} / \textcolor{blue}{26.90} & 5.7e-1 / 9.4e-1 & \textcolor{blue}{9.44} / 15.18 \\
    & 120h & 3.2e-1 / 1.9e-1 & 8.3 / \textcolor{blue}{44.5} & \textcolor{blue}{3.4e-1} / \textcolor{blue}{2e-1} & 33.8 / 40.3 & 1.9e-1 / 1.13 & 11.6 / 26.5 & 3.6e-1 / \textcolor{blue}{4.7e-1} & \textcolor{blue}{9.08} / \textcolor{blue}{26.28} & 4.9e-1 / 7.3e-1 & \textcolor{blue}{7.94} / 19.86 \\
    & 240h & 9.1e-1 / 1.1 & \textcolor{blue}{0} / 29.5 & \textcolor{blue}{5.1e-1} / \textcolor{blue}{4e-1} & \textcolor{blue}{0} / \textcolor{blue}{0} & 1.9e-1 / 9.4e-1 & 9.2 / 23.2 & \textcolor{blue}{3.6e-1} / 4.6e-1 & \textcolor{blue}{9.33} / \textcolor{blue}{27.28} & 5.1e-1 / 8.2e-1 & \textcolor{blue}{6.20} / \textcolor{blue}{17.34} \\
    \cmidrule(lr){1-12}
    \multirow{4}{*}{DLinear} & 24h & 2.6e-1 / 1.2e-1 & 17.4 / 42.2 & 2.4e-1 / 1e-1 & 38.7 / 64.2 & 3.6e-1 / 1.73 & \textcolor{blue}{31.6} / \textcolor{blue}{35} & 3.7e-1 / 4.4e-1 & 2.22 / 14.52 & 4.9e-1 / 4.9e-1 & 2.24 / 9.57 \\
    & 72h & 3.6e-1 / 2.3e-1 & 10.7 / 37.5 & 3.2e-1 / 1.8e-1 & 5.43 / 41 & 3.5e-1 / 1.44 & \textcolor{blue}{29.3} / \textcolor{blue}{33.4} & 4.1e-1 / 5.2e-1 & 1.37 / 10.80 & 6.1e-1 / \textcolor{blue}{6.9e-1} & 1.41 / 3.75 \\
    & 120h & 3.3e-1 / 1.9e-1 & 4.7 / 26.8 & 3.5e-1 / 2.1e-1 & 12.9 / 37.7 & 3.8e-1 / 1.95 & \textcolor{blue}{29.44} / \textcolor{blue}{33.78} & 4.1e-1 / 5.1e-1 & 1.14 / 10.85 & 5.3e-1 / \textcolor{blue}{5.8e-1} & 2.76 / 9.53 \\
    & 240h & 1.24 / 1.83 & \textcolor{blue}{0} / 0.88 & 7.4e-1 / 7.2e-1 & \textcolor{blue}{0} / \textcolor{blue}{0} & 3.5e-1 / 1.44 & \textcolor{blue}{29.3} / \textcolor{blue}{33.4} & 4.0e-1 / 4.9e-1 & 1.50 / 12.52 & 5.4e-1 / \textcolor{blue}{6.3e-1} & 2.33 / 9.99 \\
    \cmidrule(lr){1-12}
    \multirow{4}{*}{SparseTSF $\lozenge$} & 24h & 2.8e-1 / 1.5e-1 & 24.3 / 51.1 & 3.3e-1 / 1.8e-1 & 29.7 / 60.4 & 3.9e-1 / 2.93 & 1.1 / 4.7 & 5.4e-1 / 7.4e-1 & 0.916 / 2.60 & 7.1e-1 / 1.17 & 1.58 / 7.99 \\
    & 72h & 3.7e-1 / 2.4e-1 & \textcolor{blue}{29.3} / \textcolor{blue}{58.6} & 3.7e-1 / 2.4e-1 & 2.33 / 40.1 & 3.7e-1 / 3.61 & 0.72 / 4.40 & 5.8e-1 / 7.9e-1 & 0.507 / 2.06 & 1.00 / 1.87 & 0.490 / 2.11 \\
    & 120h & 3.3e-1 / 1.9e-1 & 10.2 / 30.8 & 4e-1 / 2.7e-1 & 18.6 / 42.2 & 3.6e-1 / 3.86 & 1.25 / 5.47 & 5.7e-1 / 7.8e-1 & 0.324 / 2.09 & 8.5e-1 / 1.50 & 0.344 / 2.56 \\
    & 240h & 1.03 / 1.21 & \textcolor{blue}{0} / 0 & 7.4e-1 / 7.3e-1 & \textcolor{blue}{0} / \textcolor{blue}{0} & 2.9e-1 / 4.82 & 0.84 / 5 & 5.7e-1 / 7.8e-1 & 0.307 / 3.47 & 8.7e-1 / 1.53 & 0.202 / 2.42 \\
    \cmidrule(lr){1-12}
    \multirow{4}{*}{TQNet \dag} & 24h & \textcolor{blue}{2.1e-1} / 8.8e-2 & 19.3 / 53.7 & 2.1e-1 / 9.1e-2 & 52.6 / 65.5 & 1.7e-1 / 0.81 & 7.36 / 19.9 & 3.1e-1 / 3.7e-1 & 6.99 / 29.9 & 3.3e-1 / 0.36 & 28.1 / 49.9 \\
    & 72h & 3.4e-1 / 2.2e-1 & \textcolor{red}{35.8} / 50.5 & 3.1e-1 / 1.8e-1 & 17.1 / \textcolor{blue}{55.6} & 2e-1 / 1.12 & 4.73 / 16.4 & 3.8e-1 / 5.0e-1 & 3.77 / 18.5 & 5.5e-1 / 0.92 & 7.62 / \textcolor{blue}{16.2} \\
    & 120h & \textcolor{blue}{3.1e-1} / \textcolor{blue}{1.8e-1} & \textcolor{red}{18.6} / 34.3 & \textcolor{blue}{3.4e-1} / 2.1e-1 & 34.6 / 39.6 & 2e-1 / 1.28 & 5.40 / 17.2 & 3.8e-1 / 5.0e-1 & 4.07 / 19.0 & 4.7e-1 / 0.72 & 7.40 / \textcolor{blue}{20.2} \\
    & 240h & 9.1e-1 / 1.06 & \textcolor{blue}{0} / 15.4 & 6.2e-1 / 5.6e-1 & \textcolor{blue}{0} / \textcolor{blue}{0} & 1.9e-1 / 0.99 & 3.92 / 15.0 & 3.7e-1 / 4.7e-1 & 4.85 / 21.7 & 4.9e-1 / 0.80 & 5.40 / 16.7 \\
    \cmidrule(lr){1-12}
    \multirow{4}{*}{\textbf{TSSM (ours)}} & 24h & \textcolor{red}{1e-1} / \textcolor{blue}{2e-2} & \textcolor{red}{32} / \textcolor{red}{80.9} & \textcolor{red}{1.7e-1} / \textcolor{red}{5e-2} & \textcolor{red}{61.69} / \textcolor{red}{78.13} & \textcolor{red}{8.6e-2} / \textcolor{red}{5.6e-1} & \textcolor{red}{45.1} / \textcolor{red}{98.4} & \textcolor{blue}{3e-1} / \textcolor{red}{2e-1} & \textcolor{red}{71.8} / \textcolor{red}{78.8} & \textcolor{red}{1e-1} / \textcolor{red}{5e-2} & \textcolor{red}{50.3} / \textcolor{red}{73} \\
    & 72h & \textcolor{red}{1.5e-1} / \textcolor{red}{4.4e-2} & 27.8 / \textcolor{red}{79.6} & \textcolor{red}{2.2e-1} / \textcolor{red}{9.4e-2} & \textcolor{red}{58.9} / \textcolor{red}{74} & \textcolor{red}{9.2e-2} / \textcolor{red}{6e-1} & \textcolor{red}{31.9} / \textcolor{red}{98} & \textcolor{red}{3.2e-1} / \textcolor{red}{3.3e-1} & \textcolor{red}{69.6} / \textcolor{red}{77.4} & \textcolor{red}{1.7e-1} / \textcolor{red}{9.2e-2} & \textcolor{red}{41.2} / \textcolor{red}{61} \\
    & 120h & \textcolor{red}{1.9e-1} / \textcolor{red}{7.1e-2} & \textcolor{blue}{12.5} / \textcolor{red}{71.4} & \textcolor{red}{2.5e-1} / \textcolor{red}{1.1e-1} & \textcolor{red}{55.7} / \textcolor{red}{72.5} & \textcolor{red}{6.1e-2} / \textcolor{red}{6e-2} & \textcolor{red}{44.7} / \textcolor{red}{98.3} & \textcolor{red}{3.4e-1} / \textcolor{red}{3.5e-1} & \textcolor{red}{69.8} / \textcolor{red}{77.2} & \textcolor{red}{2e-1} / \textcolor{red}{1.1e-1} & \textcolor{red}{36.8} / \textcolor{red}{58.1} \\
    & 240h & \textcolor{red}{2.2e-1} / \textcolor{red}{8.3e-2} & \textcolor{red}{4.81} / \textcolor{red}{52.9} & \textcolor{red}{2.9e-1} / \textcolor{red}{1.5e-1} & \textcolor{red}{50.6} / \textcolor{red}{65.2} & \textcolor{red}{6.6e-2} / \textcolor{red}{7.5e-2} & \textcolor{red}{29.4} / \textcolor{red}{98.1} & \textcolor{red}{3.4e-1} / \textcolor{red}{3.6e-1} & \textcolor{red}{70} / \textcolor{red}{76.4} & \textcolor{red}{2.3e-1} / \textcolor{red}{1.5e-2} & \textcolor{red}{21.6} / \textcolor{red}{46.6} \\
    \bottomrule
  \end{tabular*}
  \\[2pt]
  \makebox[\textwidth][l]{\small Note: $\lozenge$: period-aware AI-methods; \dag: memory-enhanced AI-methods.}
  \vspace{-10pt}
\end{table*}

\subsubsection{Quantitative Results}
We quantitatively compare the averaged accuracy and extreme weather event prediction performance of comprehensive baselines with a 48 hour look-back window for prediction horizons of 24, 48, 120, and 240 hours.

\textbf{Weather-5K Data}: As shown in Tab. \ref{tab:main_result_Weather-5K}, our method significantly outperforms the competing methods, achieving average advantages of 10\% in accuracy and 61\% in SEDI with 88\% of all metrics ranked as either the best or second-best highlighted in \textcolor{red}{red}/\textcolor{blue}{blue}. In particular, for SEDI at the 99.5th percentile, which reflects the capture capability for rare and extreme weather events, our method achieves an improvement of over 90\%. For longer forecasting windows, where atmospheric chaos becomes stronger and observational support becomes weaker, the advantage of our method is even more pronounced, reaching an average improvement of 36\% for 240h prediction. The results verify the effectiveness of history-enhanced design.

However, the period-aware and memory-enhanced methods, which are more closely related to ours, do not achieve superior performance. We attribute this to the insufficiency of reliable short-term periodic patterns and the memory confusion caused by chaotic dynamics under the short-sightedness of limited look-back windows.

Furthermore, despite a substantial disadvantage in parameter scale (5M vs. 120M), our method still overwhelmingly outperforms the fine-tuned foundation model Chronos2, achieving a 61\% overall improvement. On the one hand, this suggests that station-level weather sequences exhibit more intricate and domain-specific patterns than general time series, such that universal temporal knowledge does not necessarily translate into substantial performance gains. On the other hand, it demonstrates that the effectiveness of GSWF methods stems primarily from task-specific model design rather than scale.

Compared with operational physics-based methods \cite{ecmwf2024hres}, which assimilate massive multi-source observations through supercomputer-scale computation on dense grids, our lightweight method uses only sparse station observations yet achieves a 32.4\% average gain on 34.7\% of the metrics (25/72, marked in \textcolor{green}{green}). The gains are especially evident for transient wind variables and extreme events, highlighting TSSM's advantage in resolving local-scale dynamics and its potential as an alternative for localized weather systems.

\textbf{Hydrology and General Data}: As shown in Tab. \ref{tab:main_result_hydrology} and Tab. \ref{tab:main_result_TS}, TSSM achieves a 35\% improvement in hydrological forecasting accuracy, as well as 45\% and $>$100\% gains in accuracy and SEDI on general time series datasets, respectively. Moreover, 95\% of all metrics are ranked as either the best or second-best. These results further demonstrate the effectiveness of our approach for spatiotemporal sequences with varying degrees of human activities and energy dynamics, and reveal that empirical human behaviors and energy consumption also exhibit strong long-term evolutionary characteristics.

Notably, extreme event capture, measured by SEDI, is of critical importance in real-world scenarios, yet it remains underexplored in existing studies on general forecasting tasks. Our results show that mainstream methods exhibit limited extreme value modeling capability and often fail to capture extreme events in long-term forecasting. In contrast, TSSM maintains robust extreme event identification ability, further highlighting the importance of historical evolution modeling for reliable forecasting.

\subsubsection{Qualitative Results}

\begin{figure*}[!t]
  \centering
  \includegraphics[width=\linewidth]{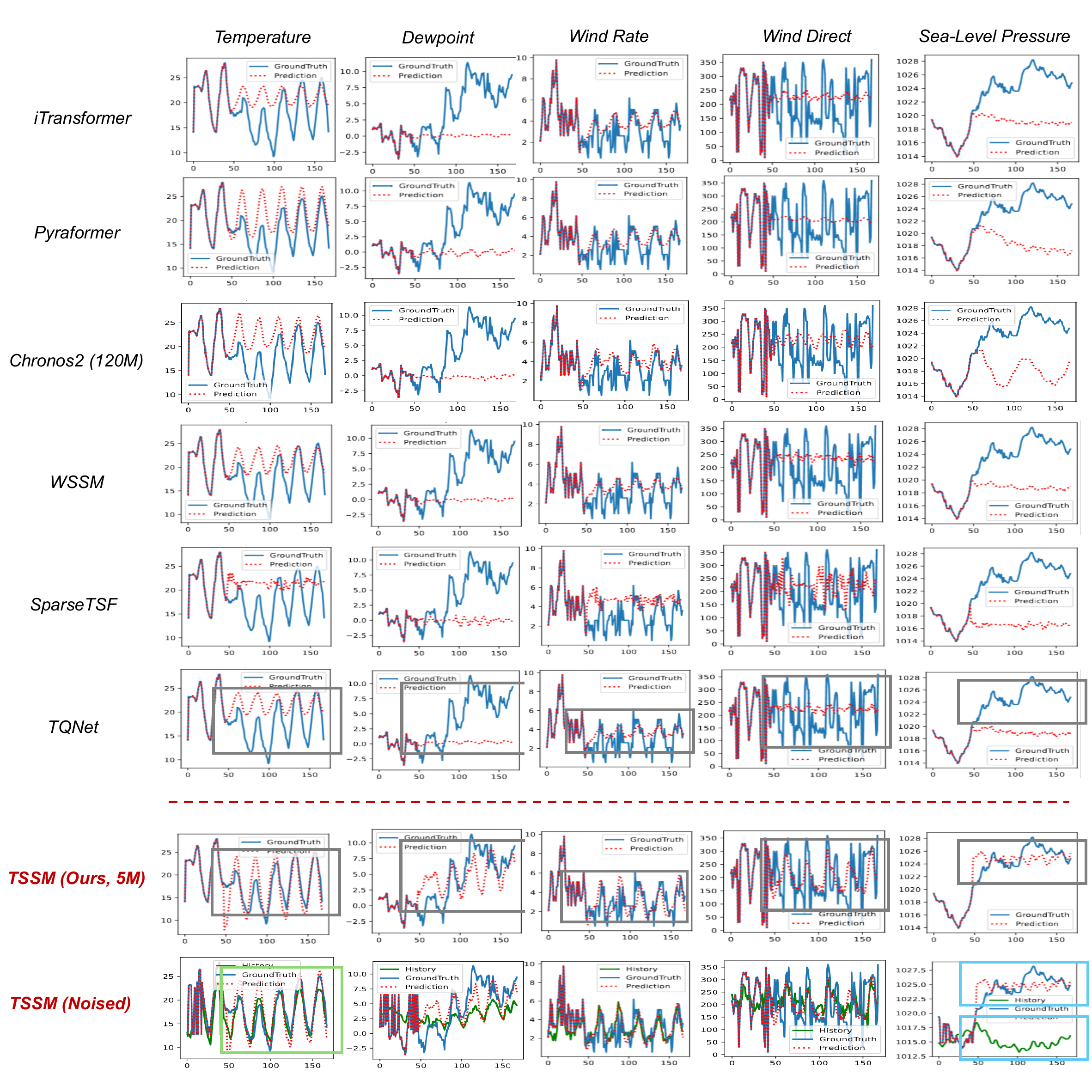}
  \caption{Visualization of 120h forecasting results across all variables, together with our corresponding results under noised look-back window. The blue solid line denotes the ground truth, the red dashed line denotes the forecast, and the green solid line denotes the history average. The gray box highlights that our method outperforms existing methods, while the green and blue boxes demonstrate our effective utilization of historical information.}
  \label{fig:vis}
\end{figure*}

To verify the reliability of the qualitative results, we further visualize the GSWF outputs on challenging samples across all variables, as illustrated in Fig. \ref{fig:vis}. It can be observed in gray boxes that our method substantially improves prediction accuracy and yields high-frequency patterns that are more consistent with extreme values. Meanwhile, we visualized history under noisy conditions, as indicated by the green box. Historically robust patterns that cover both look-back and forecasting window provide a strong reference and help correct noise contamination. More importantly, as shown in the blue boxes, our method does not blindly mimic history: even when historical information is misleading, it can still identify and output the correct forecast. In contrast, baselines tend to only capture low-frequency patterns and collapse toward mean values. This limitation stems from insufficient observations within limited look-back windows to support long-range prediction, which also explains the poor SEDI performance of existing approaches.

In summary, this qualitative comparison further demonstrates the effectiveness of TSSM and highlights the supportive role of historical information in forecasting, which validates the rationality of our design insight.

\vspace{-10pt}
\subsection{Ensemble Forecasting}
By producing multiple candidate predictions supported by different historical observation anchors, TSSM naturally supports ensemble forecasting without relying on complex generative frameworks.

We compare deterministic methods, which add noise to features before decoding to generate diverse predictions, with probabilistic generative methods that natively support ensemble forecasting. Since the number of past years is $H-1=9$, we set the number of candidate predictions to $k=9$.

As shown in Tab. \ref{tab:result_ensemble}, our method substantially improves ensemble prediction accuracy, even surpassing the advanced probabilistic method. However, because the Best-in-9 results are selected according to accuracy, no obvious gain is observed in SEDI. This result demonstrates that multi-year aligned historical data can serve as effective anchors for ensemble forecasting. In contrast, the other deterministic baselines show no evident improvement, indicating that simply increasing the number of candidate forecasts is insufficient to improve ensemble prediction performance.

\begin{table*}[!t]
  \centering
  \caption{Accuracy and Extreme Event Capture Comparison on \textbf{WEATHER DATA} for \textbf{Ensemble Forecasting} with best in $k=9$ \vspace{-5pt}}
  \label{tab:result_ensemble}
  \scriptsize
  \setlength{\tabcolsep}{2pt}
  \renewcommand{\arraystretch}{0.82}
  \begin{tabular*}{\textwidth}{@{\extracolsep{\fill}}@{}p{5.2em}@{}c@{}ccccccccc}
    \toprule
    \multirow{2}{*}{Baselines} & \multirow{2}{*}{Window} &
    \multicolumn{2}{c}{Temperature} & \multicolumn{2}{c}{Dewpoint} & \multicolumn{2}{c}{Wind Rate} & \multicolumn{1}{c}{Wind Direc.} & \multicolumn{2}{c}{Sea-Level Pressure} \\
    \cmidrule(lr){3-4} \cmidrule(lr){5-6} \cmidrule(lr){7-8} \cmidrule(lr){9-9} \cmidrule(lr){10-11}
    & & MAE/MSE$\downarrow$ & 99.5th/90.0th$\uparrow$ & MAE/MSE$\downarrow$ & 99.5th/90.0th$\uparrow$ & MAE/MSE$\downarrow$ & 99.5th/90.0th$\uparrow$ & MAE/MSE($\times 10^3$)$\downarrow$ & MAE/MSE$\downarrow$ & 99.5th/90.0th$\uparrow$\\
    \midrule
    \multirow{1}{*}{Pyraformer}
    & 120h & 3.05 / 18.3 & 6.03 / 42.7 & 2.93 / 19.2 & 4.75 / 34.5 & 1.41 / 4.74 & 1.29 / 7.11 & 73.5 / 8.07 & 4.47 / 40.9 & 5.85 / 19.2 \\
    \cmidrule(lr){1-11}
    \multirow{1}{*}{Dlinear}
    & 120h & 3.77 / 25.1 & 2.88 / 19.4 & 3.81 / 25.0 & 3.32 / 21.0 & 1.59 / 5.97 & 0.33 / 7.21 & 74.1 / 8.71 & 4.81 / 49.0 & 1.94 / 17.4 \\
    \cmidrule(lr){1-11}
    \multirow{1}{*}{WSSM}
    & 120h & 2.94 / 17.0 & 17.6 / 56.5 & 3.04 / 22.0 & 12.5 / 53.5 & 1.55 / 5.63 & 5.32 / 14.3 & 69.1 / 7.12 & 4.84 / 52.9 & 12.9 / 45.1 \\
    \cmidrule(lr){1-11}
    \multirow{1}{*}{Diff-TS~\cite{yuan2024diffusionts}}
    & 120h & 2.57 / 13.15 & 34.6 / 67.5 & 2.87 / 17.9 & 23.5 / 58.3 & 1.57 / 5.33 & 10 / 23.6 & 75.2 / 9.82 & 4.59 / 45.3 & 24.9 / 49.1 \\
    \cmidrule(lr){1-11}
    \multirow{1}{*}{\textbf{TSSM (ours)}}
    & 120h & 1.71 / 6.99 & 34.1 / 71.3 & 1.92 / 8.49 & 12.9 / 61.5 & 1.02 / 2.40 & 13.3 / 24.3 & 52.2 / 5.35 & 2.80 / 18.0 & 23.7 / 49.3 \\
    \bottomrule
  \end{tabular*}
  \vspace{-10pt}
\end{table*}

\subsection{Long-Horizon Forecasting}
For long-horizon forecasting, existing paradigms rely primarily on short-term observations and are therefore unable to perform reliable one-step long-horizon forecasting. Moreover, during iterative long-term prediction, subsequent forecasts can only be based on inaccurate predictions from the previous step, leading to severe error accumulation and pattern drift.

By introducing the historical dimension, our method not only utilizes historical information to achieve stronger performance in one-step long-horizon forecasting, but also calibrates errors at each step with sufficient historical support, thereby greatly alleviating error accumulation and pattern drift in iterative forecasting.

As shown in Tab. \ref{tab:result_long}, TSSM not only achieves 37.5\% substantial improvements in 240h long-horizon forecasting, but also shows that iterative short-term forecasting, such as 120h $\times$ 2 and 48h $\times$ 5, outperforms direct 240h $\times$ 1 long-horizon forecasting, reaching a 103.5\% gain. Owing to limited error accumulation, compared with direct 120h forecasting, iterative 120h $\times$ 2 forecasting leads to only a marginal 4.3\% degradation in accuracy and SEDI. This property is particularly valuable in weather forecasting, where extending the effective forecasting window is highly desirable.

\begin{table*}[!t]
  \centering
  \caption{Accuracy and Extreme Event Capture Comparison on \textbf{WEATHER DATA} for \textbf{Long Time (10 days/240h) Forecasting} \vspace{-5pt}}
  \label{tab:result_long}
  \scriptsize
  \setlength{\tabcolsep}{2pt}
  \renewcommand{\arraystretch}{0.82}
  \begin{tabular*}{\textwidth}{@{\extracolsep{\fill}}@{}c@{}p{5.2em}@{}c@{}ccccccccc}
    \toprule
    & \multirow{2}{*}{Baselines} & \multirow{2}{*}{Window} &
    \multicolumn{2}{c}{Temperature} & \multicolumn{2}{c}{Dewpoint} & \multicolumn{2}{c}{Wind Rate} & \multicolumn{1}{c}{Wind Direc.} & \multicolumn{2}{c}{Sea-Level Pressure} \\
    \cmidrule(lr){4-5} \cmidrule(lr){6-7} \cmidrule(lr){8-9} \cmidrule(lr){10-10} \cmidrule(lr){11-12}
    & & & MAE/MSE$\downarrow$ & 99.5th/90.0th$\uparrow$ & MAE/MSE$\downarrow$ & 99.5th/90.0th$\uparrow$ & MAE/MSE$\downarrow$ & 99.5th/90.0th$\uparrow$ & MAE/MSE($\times 10^3$)$\downarrow$ & MAE/MSE$\downarrow$ & 99.5th/90.0th$\uparrow$\\
    \midrule
    \multicolumn{2}{@{}c@{}}{WSSM} & 240h & 3.44 / 23 & 14.7 / 55.3 & 3.6 / 25.9 & 12.9 / 42.1 & 1.62 / 5.57 & 4.89 / 18.1 & 78.9 / 9.08 & 5.57 / 63.9 & 8.91 / 31.2 \\
    \midrule
    \multirow{3}{*}{\raisebox{-\height}{\rotatebox{90}{TSSM}}}
    & 240h $\times$ 1 & 240h & 3.04 / 17.2 & 47 / 74.5 & 2.83 / 16.2 & 6.4 / 45.6 & 1.49 / 4.4 & 13.9 / 24.6 & 70.5 / 8.43 & 3.69 / 28.1 & 8.26 / 36.5 \\
    \cmidrule(lr){2-12}
    & 120h $\times$ 2 & 240h & 2.69 / 15.3 & 45.4 / 72.8 & 2.57 / 14.1 & 13.6 / 54.4 & 1.46 / 4.54 & 10.8 / 21.0 & 68.4 / 8.89 & 3.85 / 29.4 & 11.0 / 43.6 \\
    \cmidrule(lr){2-12}
    & 48h $\times$ 5 & 240h & 2.47 / 12.8 & 62.6 / 80.6 & 1.83 / 7.56 & 25 / 68.2 & 1.3 / 3.74 & 20.7 / 33.1 & 60.2 / 6.87 & 2.19 / 11.6 & 42 / 69.4 \\
    \bottomrule
  \end{tabular*}
  \vspace{-10pt}
\end{table*}

\begin{figure}[!t]
  \centering
  \includegraphics[width=\linewidth]{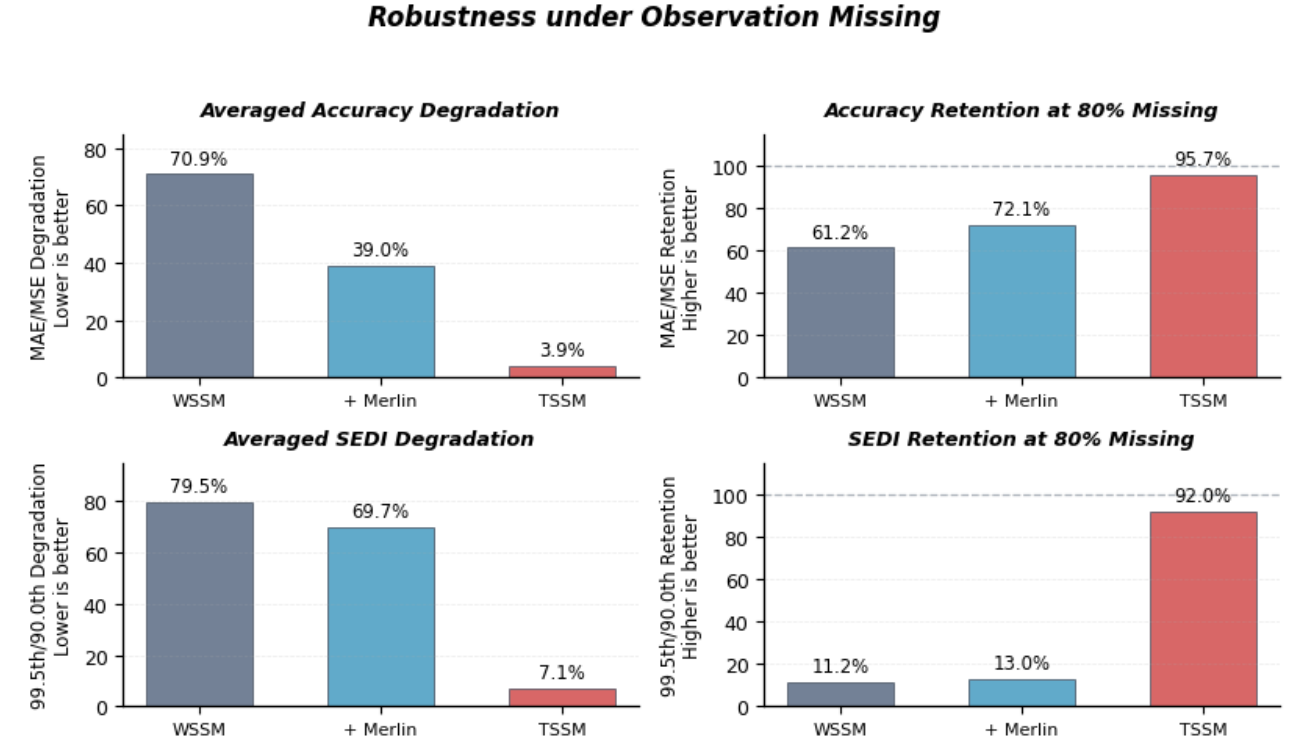}
  \caption{Robustness comparison under observation missingness. The left panels show the averaged degradation over 20\%, 40\%, and 80\% missing ratios, where lower is better; the right panels show the retention at 80\% missing, where higher is better. TSSM (marked in red) exhibits the lowest degradation and the highest retention, indicating strong robustness to observation missingness.}
  \label{fig:vis_noise}
\end{figure}

\subsection{Forecasting under Observation Missing}

\begin{table*}[!t]
  \centering
  \caption{Accuracy and Extreme Event Capture Comparison on \textbf{WEATHER DATA} under \textbf{OBSERVATION MISSINGNESS} \vspace{-5pt}}
  \label{tab:result_missing}
  \scriptsize
  \setlength{\tabcolsep}{2pt}
  \renewcommand{\arraystretch}{0.82}
  \begin{tabular*}{\textwidth}{@{\extracolsep{\fill}}@{}p{5.2em}@{}c@{}ccccccccc}
    \toprule
    \multirow{2}{*}{Baselines} & \multirow{2}{*}{Missing} &
    \multicolumn{2}{c}{Temperature} & \multicolumn{2}{c}{Dewpoint} & \multicolumn{2}{c}{Wind Rate} & \multicolumn{1}{c}{Wind Direc.} & \multicolumn{2}{c}{Sea-Level Pressure} \\
    \cmidrule(lr){3-4} \cmidrule(lr){5-6} \cmidrule(lr){7-8} \cmidrule(lr){9-9} \cmidrule(lr){10-11}
    & & MAE/MSE$\downarrow$ & 99.5th/90.0th$\uparrow$ & MAE/MSE$\downarrow$ & 99.5th/90.0th$\uparrow$ & MAE/MSE$\downarrow$ & 99.5th/90.0th$\uparrow$ & MAE/MSE($\times 10^3$)$\downarrow$ & MAE/MSE$\downarrow$ & 99.5th/90.0th$\uparrow$\\
    \midrule
    \multirow{3}{*}{WSSM}
    & 0\% & 2.97 / 17.8 & 18.22 / 58.2 & 3.08 / 19.7 & 13.57 / 45.2 & 1.59 / 5.91 & 4.98 / 12.4 & 75.8 / 8.78 & 4.83 / 54.0 & 10.5 / 37.9 \\
    \cmidrule(lr){2-11}
    & 20\% & 4.08 / 29.0 & 2.38 / 22.6 & 4.07 / 30.1 & 2.80 / 22.5 & 1.70 / 5.98 & 0.42 / 5.06 & 78.6 / 9.68 & 5.53 / 62.1 & 2.07 / 19.7 \\
    & 40\% & 5.22 / 46.5 & 1.42 / 12.4 & 5.02 / 43.0 & 1.68 / 12.1 & 1.73 / 6.00 & 0.20 / 3.51 & 79.1 / 9.33 & 5.56 / 60.3 & 3.55 / 9.62 \\
    & 80\% & 8.42 / 116 & 1.62 / 7.74 & 7.81 / 101 & 1.37 / 6.98 & 1.92 / 7.12 & 0.01 / 3.33 & 82.5 / 9.43 & 6.30 / 72.3 & 0.93 / 2.31 \\
    \midrule
    \multirow{3}{*}{ + Merlin~\cite{yu2025merlin}}
    & 20\% & 3.72 / 25.9 & 3.65 / 32.3 & 3.76 / 27.4 & 5.12 / 32.3 & 1.70 / 6.11 & 0.75 / 7.92 & 78.7 / 9.61 & 5.62 / 65.2 & 4.13 / 25.7 \\
    & 40\% & 4.59 / 36.8 & 1.82 / 16.5 & 4.49 / 35.3 & 2.07 / 16.3 & 1.71 / 5.94 & 0.31 / 4.44 & 78.7 / 9.47 & 5.51 / 60.6 & 8.82 / 14.2 \\
    & 80\% & 5.94 / 59.5 & 1.44 / 10.2 & 5.64 / 53.6 & 1.49 / 9.71 & 1.76 / 6.14 & 0.13 / 3.26 & 79.7 / 9.25 & 5.67 / 61.4 & 0.12 / 6.08 \\
    \midrule
    \multirow{3}{*}{TSSM}
    & 0\% & 2.54 / 13.7 & 45.6 / 73.3 & 2.51 / 13.2 & 16.4 / 57.5 & 1.44 / 4.46 & 9.58 / 22.5 & 68.7 / 8.04 & 3.72 / 28.6 & 12.3 / 44.8 \\
    \cmidrule(lr){2-11}
    & 20\% & 2.63 / 14.9 & 44.2 / 73.1 & 2.59 / 14.5 & 12.2 / 57.3 & 1.42 / 4.58 & 7.49 / 24.7 & 69.2 / 8.03 & 3.88 / 29.2 & 11.3 / 43.9 \\
    & 40\% & 2.64 / 15.0 & 44.1 / 73.0 & 2.60 / 14.6 & 12.2 / 57.2 & 1.42 / 4.58 & 6.97 / 24.6 & 69.3 / 8.06 & 3.90 / 29.5 & 11.6 / 43.8 \\
    & 80\% & 2.64 / 15.1 & 43.8 / 72.9 & 2.61 / 14.8 & 11.8 / 57.0 & 1.43 / 4.62 & 6.88 / 24.5 & 69.5 / 8.09 & 3.94 / 30.1 & 11.3 / 43.5 \\
    \bottomrule
  \end{tabular*}
  \vspace{-10pt}
\end{table*}

Due to sensor failure, transmission packet loss, et al., global weather stations suffer from widespread missing observations. Existing methods typically employ an independent preprocessing step to impute missing values before model forecast, which introduces additional systematic and cascading errors.

By introducing the historical dimension, our method can compensate for the missing temporal observations using period-aligned information. Therefore, we take a step further to perform forecasting directly from incomplete observations.

As shown in Fig. \ref{fig:vis_noise} and Tab. \ref{tab:result_missing}, when exposed to varying degrees (20\% to 80\%) of random data missing, TSSM exhibits strong robustness against missing observations in the look-back window with only averaged 4.3\% and 8\% degradation on accuracy and SEDI. Even with 80\% of observations missing, it can still retain 95.7\% and 92\% of its performance on accuracy and SEDI by relying on historical support. In contrast, for baseline methods, introducing missing noise severely degrades performance. Although state-of-the-art noise-robust forecasting paradigms can significantly alleviate this issue, their performance still drops considerably.

Notably, since we employ historical data, missing observations in historical records from at least one year ago have been sufficiently recovered. Therefore, we do not introduce observation missing in the historical dimension.

\subsection{Forecasting under Limited Data}

\begin{table*}[!t]
  \centering
  \caption{Accuracy and Extreme Event Capture Comparison on \textbf{Limited WEATHER DATA} in \textbf{Africa and South America} \vspace{-5pt}}
  \label{tab:result_few}
  \scriptsize
  \setlength{\tabcolsep}{2pt}
  \renewcommand{\arraystretch}{0.82}
  \begin{tabular*}{\textwidth}{@{\extracolsep{\fill}}@{}p{5.2em}@{}c@{}ccccccccc}
    \toprule
    \multirow{2}{*}{Baselines} & \multirow{2}{*}{Window} &
    \multicolumn{2}{c}{Temperature} & \multicolumn{2}{c}{Dewpoint} & \multicolumn{2}{c}{Wind Rate} & \multicolumn{1}{c}{Wind Direc.} & \multicolumn{2}{c}{Sea-Level Pressure} \\
    \cmidrule(lr){3-4} \cmidrule(lr){5-6} \cmidrule(lr){7-8} \cmidrule(lr){9-9} \cmidrule(lr){10-11}
    & & MAE/MSE$\downarrow$ & 99.5th/90.0th$\uparrow$ & MAE/MSE$\downarrow$ & 99.5th/90.0th$\uparrow$ & MAE/MSE$\downarrow$ & 99.5th/90.0th$\uparrow$ & MAE/MSE($\times 10^3$)$\downarrow$ & MAE/MSE$\downarrow$ & 99.5th/90.0th$\uparrow$\\
    \midrule
    \multirow{1}{*}{WSSM AFR} & 120h & 3.82 / 27.3 & 11.1 / 49.0 & 3.68 / 28.5 & 9.2 / 40.5 & 1.71 / 6.21 & 0.99 / 9.99 & 79.3 / 10.1 & 5.54 / 65.1 & 9.77 / 33.4 \\
    \cmidrule(lr){1-11}
    \multirow{1}{*}{WSSM SAM} & 120h & 4.29 / 32.5 & 6.81 / 35.4 & 3.98 / 31.0 & 4.19 / 36.8 & 1.77 / 6.33 & 0.37 / 6.71 & 79.6 / 9.88 & 5.66 / 63.5 & 6.25 / 22.4 \\
    \cmidrule(lr){1-11}
    \multirow{1}{*}{TSSM AFR} & 120h & 2.56 / 14.1 & 66.0 / 82.6 & 2.34 / 12.1 & 16.8 / 51.6 & 1.54 / 5.34 & 16.1 / 29.9 & 70.9 / 9.04 & 2.06 / 9.49 & 23.1 / 56.5 \\
    \cmidrule(lr){1-11}
    \multirow{1}{*}{TSSM SAM} & 120h & 2.76 / 17.3 & 65.0 / 78.4 & 2.00 / 10.4 & 7.62 / 46.4 & 1.57 / 5.29 & 13.0 / 30.4 & 65.6 / 8.18 & 2.64 / 16.2 & 21.3 / 51.2 \\
    \bottomrule
  \end{tabular*}
  \vspace{-10pt}
\end{table*}

The rapid accumulation of station weather data does not guarantee that sufficient data are naturally available across all regions. Data imbalance remains severe, requiring models to adapt to limited-data scenarios.

Although station observations provide global coverage, weather stations remain notably sparse in regions such as Africa (AFR) and South America (SAM) due to uneven regional development. Despite accounting for approximately 30\% of global land area and population, these two regions host only 331 and 123 stations, respectively, together representing merely 1/14 of global observations.

As shown in Tab. \ref{tab:result_few}, when trained on only a very small number of stations, the advantage of our method over baseline approaches becomes even more significant, with improvements of 38\% in average accuracy and $>$100\% in extreme value capture. By contrast, the competing methods can hardly learn weather dynamics from limited data.

Benefiting from strong support from historical information, TSSM can model weather dynamics more effectively, leading to superior performance when trained from scratch under limited-sample conditions in these two regions.

\subsection{Ablation Study}

\begin{table*}[!t]
  \centering
  \caption{Accuracy and Extreme Event Capture Comparison on \textbf{WEATHER DATA} with \textbf{Varying Model Structure} \vspace{-5pt}}
  \label{tab:result_struct}
  \scriptsize
  \setlength{\tabcolsep}{2pt}
  \renewcommand{\arraystretch}{0.82}
  \begin{tabular*}{\textwidth}{@{\extracolsep{\fill}}@{}p{5.2em}@{}c@{}ccccccccc}
    \toprule
    \multirow{2}{*}{Baselines} & \multirow{2}{*}{Window} &
    \multicolumn{2}{c}{Temperature} & \multicolumn{2}{c}{Dewpoint} & \multicolumn{2}{c}{Wind Rate} & \multicolumn{1}{c}{Wind Direc.} & \multicolumn{2}{c}{Sea-Level Pressure} \\
    \cmidrule(lr){3-4} \cmidrule(lr){5-6} \cmidrule(lr){7-8} \cmidrule(lr){9-9} \cmidrule(lr){10-11}
    & & MAE/MSE$\downarrow$ & 99.5th/90.0th$\uparrow$ & MAE/MSE$\downarrow$ & 99.5th/90.0th$\uparrow$ & MAE/MSE$\downarrow$ & 99.5th/90.0th$\uparrow$ & MAE/MSE($\times 10^3$)$\downarrow$ & MAE/MSE$\downarrow$ & 99.5th/90.0th$\uparrow$\\
    \midrule
    \multirow{1}{*}{TSSM} & 120h & 2.54 / 13.7 & 45.6 / 73.3 & 2.51 / 13.2 & 16.4 / 57.5 & 1.44 / 4.46 & 9.58 / 22.5 & 68.7 / 8.04 & 3.72 / 28.6 & 12.3 / 44.8 \\
    \cmidrule(lr){1-11}
    \multirow{1}{*}{No C-shuffle} & 120h & 2.75 / 16.0 & 41.2 / 70.8 & 2.60 / 14.7 & 13.4 / 55.1 & 1.44 / 4.64 & 8.13 / 20.5 & 72.1 / 9.94 & 3.88 / 29.7 & 11.9 / 43.8 \\
    \cmidrule(lr){1-11}
    \multirow{1}{*}{1-Level} & 120h & 2.39 / 11.6 & 32.1 / 55.0 & 2.37 / 12.0 & 9.73 / 37.7 & 1.34 / 3.51 & 7.26 / 20.4 & 66.6 / 7.74 & 3.34 / 24.5 & 11.0 / 39.8 \\
    \cmidrule(lr){1-11}
    \multirow{1}{*}{3-Level} & 120h & 3.49 / 22.9 & 71.0 / 81.5 & 2.52 / 14.0 & 19.6 / 58.4 & 1.49 / 4.4 & 14.8 / 32.4 & 72.8 / 9.43 & 3.83 / 31.3 & 21.1 / 50.3 \\
    \cmidrule(lr){1-11}
    \multirow{1}{*}{Avg-TSSM} & 120h & 2.09 / 9.77 & 13.3 / 60.6 & 2.38 / 11.88 & 7.1 / 42.9 & 1.7 / 4.37 & 3.97 / 12.75 & 65.41 / 7.55 & 3.2 / 19.7 & 6.71 / 34.7 \\
    \cmidrule(lr){1-11}
    \multirow{1}{*}{Avg-iTrans} & 120h & 9.56 / 147.3 & 69.7 / 79.5 & 8.89 / 131.4 & 60.1 / 74.6 & 1.97 / 7.01 & 48.1 / 42.6 & 78.35 / 10.03 & 6.46 / 73.6 & 27.6 / 50.8 \\
    \cmidrule(lr){1-11}
    \multirow{1}{*}{T+H} & 120h & 3.09 / 17.6 & 63.5 / 78.9 & 2.81 / 17.1 & 19.3 / 54.3 & 1.56 / 4.91 & 19.2 / 34.9 & 76.1 / 10.2 & 3.98 / 35.7 & 19.5 / 50.2 \\
    \cmidrule(lr){1-11}
    \multirow{1}{*}{T+C} & 120h & 2.01 / 8 & 16.2 / 64.4 & 2.23 / 10.01 & 8.33 / 50.6 & 1.26 / 3.17 & 3.01 / 10.2 & 63.7 / 6.83 & 2.84 / 16.8 & 9.28 / 44.4 \\
    \cmidrule(lr){1-11}
    \multirow{1}{*}{H+C} & 120h & 3.03 / 17.3 & 61.3 / 78.6 & 2.68 / 15.6 & 16.6 / 54.3 & 1.53 / 4.69 & 17.3 / 32.4 & 73.4 / 9.61 & 3.88 / 35.6 & 16.6 / 49.5 \\
    \cmidrule(lr){1-11}
    \multirow{1}{*}{H} & 120h & 3.39 / 21.0 & 65.0 / 79.7 & 2.79 / 16.9 & 19.4 / 54.7 & 1.59 / 5.02 & 19.8 / 34.9 & 76.3 / 10.35 & 3.95 / 37.2 & 17.3 / 49.8 \\
    \cmidrule(lr){1-11}
    \multirow{1}{*}{T} & 120h & 2.05 / 8.04 & 8.62 / 58.9 & 2.24 / 10.17 & 10.0 / 51.2 & 1.27 / 3.17 & 3.74 / 11.1 & 63.6 / 6.91 & 2.84 / 16.7 & 7.37 / 43.5 \\
    \bottomrule
  \end{tabular*}
  \vspace{-10pt}
\end{table*}

\begin{table*}[!t]
  \centering
  \caption{Accuracy and Extreme Event Capture Comparison on \textbf{WEATHER DATA} with \textbf{Historical Scope $h_{st}$ $\rightarrow$ 2021} \vspace{-5pt}}
  \label{tab:main_result_hist}
  \scriptsize
  \setlength{\tabcolsep}{2pt}
  \renewcommand{\arraystretch}{0.82}
  \begin{tabular*}{\textwidth}{@{\extracolsep{\fill}}@{}p{5.2em}@{}c@{}ccccccccc}
    \toprule
    \multirow{2}{*}{Baselines} & \multirow{2}{*}{Window} &
    \multicolumn{2}{c}{Temperature} & \multicolumn{2}{c}{Dewpoint} & \multicolumn{2}{c}{Wind Rate} & \multicolumn{1}{c}{Wind Direc.} & \multicolumn{2}{c}{Sea-Level Pressure} \\
    \cmidrule(lr){3-4} \cmidrule(lr){5-6} \cmidrule(lr){7-8} \cmidrule(lr){9-9} \cmidrule(lr){10-11}
    & & MAE/MSE$\downarrow$ & 99.5th/90.0th$\uparrow$ & MAE/MSE$\downarrow$ & 99.5th/90.0th$\uparrow$ & MAE/MSE$\downarrow$ & 99.5th/90.0th$\uparrow$ & MAE/MSE($\times 10^3$)$\downarrow$ & MAE/MSE$\downarrow$ & 99.5th/90.0th$\uparrow$\\
    \midrule
    \multirow{1}{*}{$h_{st}$=2014} & 120h & 2.54 / 13.7 & 45.6 / 73.3 & 2.51 / 13.2 & 16.4 / 57.5 & 1.44 / 4.46 & 9.58 / 22.5 & 68.7 / 8.04 & 3.72 / 28.6 & 12.3 / 44.8 \\
    \cmidrule(lr){1-11}
    \multirow{1}{*}{$h_{st}$=2015} & 120h & 2.53 / 11.8 & 43.1 / 73 & 2.4 / 11.9 & 12.8 / 53.4 & 1.41 / 3.91 & 8.44 / 21.2 & 63.8 / 6.99 & 3.49 / 22.7 & 12 / 44.5 \\
    \cmidrule(lr){1-11}
    \multirow{1}{*}{$h_{st}$=2016} & 120h & 2.31 / 10.5 & 39.6 / 71.6 & 2.27 / 10.6 & 10.1 / 52.1 & 1.29 / 3.41 & 7.5 / 18.9 & 64.5 / 7.23 & 3.29 / 18.9 & 11.7 / 43.9 \\
    \cmidrule(lr){1-11}
    \multirow{1}{*}{$h_{st}$=2017} & 120h & 2.15 / 8.93 & 21.0 / 64.8 & 2.28 / 10.6 & 11.7 / 52.1 & 1.30 / 3.42 & 6.11 / 16.2 & 65.2 / 7.34 & 3.14 / 17.2 & 10.4 / 43.1 \\
    \cmidrule(lr){1-11}
    \multirow{1}{*}{$h_{st}$=2018} & 120h & 2.07 / 8.56 & 32.7 / 70.2 & 2.24 / 10.3 & 9.96 / 52.0 & 1.26 / 3.25 & 5.1 / 17.2 & 63.4 / 6.96 & 2.91 / 17.2 & 11.1 / 43.3 \\
    \cmidrule(lr){1-11}
    \multirow{1}{*}{$h_{st}$=2019} & 120h & 2.04 / 8.28 & 28.7 / 69.0 & 2.25 / 10.3 & 9.93 / 51.8 & 1.26 / 3.22 & 4.02 / 16.0 & 63.6 / 7.00 & 2.86 / 17.2 & 10.7 / 43.4 \\
    \cmidrule(lr){1-11}
    \multirow{1}{*}{$h_{st}$=2020} & 120h & 2.03 / 8.00 & 14.8 / 62.4 & 2.23 / 10.1 & 8.59 / 51.3 & 1.26 / 3.18 & 3.42 / 12.2 & 63.4 / 6.92 & 2.84 / 16.8 & 9.69 / 42.5 \\
    \bottomrule
  \end{tabular*}
  \vspace{-10pt}
\end{table*}

\begin{table*}[!t]
  \centering
  \caption{Accuracy and Extreme Event Capture Comparison on \textbf{120h WEATHER DATA} with \textbf{History Stacked in 8$\times$48h look-back windows} \vspace{-5pt}}
  \label{tab:main_result_time}
  \scriptsize
  \setlength{\tabcolsep}{2pt}
  \renewcommand{\arraystretch}{0.82}
  \begin{tabular*}{\textwidth}{@{\extracolsep{\fill}}@{}p{5.2em}@{}c@{}ccccccccc}
    \toprule
    \multirow{2}{*}{Baselines} & \multirow{2}{*}{Look-back} &
    \multicolumn{2}{c}{Temperature} & \multicolumn{2}{c}{Dewpoint} & \multicolumn{2}{c}{Wind Rate} & \multicolumn{1}{c}{Wind Direc.} & \multicolumn{2}{c}{Sea-Level Pressure} \\
    \cmidrule(lr){3-4} \cmidrule(lr){5-6} \cmidrule(lr){7-8} \cmidrule(lr){9-9} \cmidrule(lr){10-11}
    & & MAE/MSE$\downarrow$ & 99.5th/90.0th$\uparrow$ & MAE/MSE$\downarrow$ & 99.5th/90.0th$\uparrow$ & MAE/MSE$\downarrow$ & 99.5th/90.0th$\uparrow$ & MAE/MSE($\times 10^3$)$\downarrow$ & MAE/MSE$\downarrow$ & 99.5th/90.0th$\uparrow$\\
    \midrule
    \multirow{2}{*}{iTransformer}
    & 48h & 3.09 / 21.2 & 14.6 / 52.1 & 3.34 / 24.9 & 9.78 / 44.7 & 1.7 / 6.17 & 1.42 / 10.3 & 77.8 / 9.64 & 5.18 / 58.5 & 13 / 36.6 \\
    \cmidrule(lr){2-11}
    & 384h & 2.92 / 20.0 & 12.18 / 46.1 & 3.37 / 22.7 & 4.49 / 36.2 & 1.57 / 5.26 & 1.42 / 8.29 & 75.5 / 8.92 & 5.05 / 52.4 & 4.92 / 20.2 \\
    \cmidrule(lr){1-11}
    \multirow{2}{*}{Chronos2}
    & 48h & 3.32 / 23.3 & 16.8 / 27.2 & 3.52 / 29.5 & 10.3 / 19.6 & 1.64 / 5.53 & 4.48 / 6.03 & 78.4 / 10.1 & 4.95 / 54.9 & 11.9 / 17.6 \\
    \cmidrule(lr){2-11}
    & 384h & 3.08 / 19.8 & 13.7 / 54.2 & 3.34 / 23.8 & 9.52 / 43.1 & 1.54 / 5.37 & 4.01 / 12.9 & 74.3 / 9.65 & 4.43 / 43.1 & 12.1 / 32.1 \\
    \cmidrule(lr){1-11}
    \multirow{2}{*}{WSSM}
    & 48h & 2.97 / 17.8 & 18.22 / 58.2 & 3.08 / 19.7 & 13.57 / 45.2 & 1.59 / 5.91 & 4.98 / 12.4 & 75.8 / 8.78 & 4.83 / 54.0 & 10.5 / 37.9 \\
    \cmidrule(lr){2-11}
    & 384h & 3.02 / 17.8 & 19.8 / 60.2 & 3.27 / 24.7 & 11.2 / 49.5 & 1.67 / 5.96 & 7.25 / 19.3 & 75.1 / 9.25 & 4.94 / 53 & 11.8 / 43.9 \\
    \cmidrule(lr){1-11}
    \multirow{2}{*}{SparseTSF}
    & 48h & 4.15 / 31.2 & 4.8 / 34 & 3.8 / 29.2 & 9.1 / 43.2 & 1.82 / 6.93 & 0.648 / 7.84 & 82.1 / 10.7 & 6.16 / 75.8 & 4.76 / 28.6 \\
    \cmidrule(lr){2-11}
    & 384h & 3.45 / 22.6 & 0.77 / 39.5 & 3.61 / 25.3 & 1.39 / 31.7 & 1.62 / 5.53 & 0.68 / 6.18 & 77.4 / 9.19 & 5.49 / 59.6 & 3.05 / 13.2 \\
    \cmidrule(lr){1-11}
    \multirow{2}{*}{TQNet}
    & 48h & 3.55 / 24.6 & 7.03 / 46.5 & 3.59 / 27 & 11.4 / 47.2 & 1.71 / 6.2 & 1.03 / 9.59 & 79.2 / 10.2 & 5.54 / 65 & 8.79 / 32.8 \\
    \cmidrule(lr){2-11}
    & 384h & 3.18 / 19.9 & 4.10 / 47.0 & 3.37 / 22.8 & 3.78 / 36.2 & 1.57 / 5.24 & 1.16 / 8.08 & 75.5 / 8.96 & 5.09 / 53.2 & 3.45 / 19.2 \\
    \cmidrule(lr){1-11}
    \multirow{1}{*}{TSSM}
    & 48h & 2.54 / 13.7 & 45.6 / 73.3 & 2.51 / 13.2 & 16.4 / 57.5 & 1.44 / 4.46 & 9.58 / 22.5 & 68.7 / 8.04 & 3.72 / 28.6 & 12.3 / 44.8 \\
    \bottomrule
  \end{tabular*}
  \\[2pt]
  \makebox[\textwidth][l]{\small Note: TSSM is included here as a reference with explicitly arranged 8-year historical axis; therefore, has only 48h look-back window.}
  \vspace{-10pt}
\end{table*}

\subsubsection{Model Structure Analysis}
As shown in Tab. \ref{tab:result_struct}, we permute and combine the three branches (T-Scan, V-Scan, H-Scan) while retaining at least one of the temporal and historical branches. When the temporal branch is removed, information from the look-back window is injected into the prediction window through modulation. We also ablate several other design choices, including the hierarchical structure level and variable-dimension shuffling. Although our default setting does not explicitly use historical observations, we further explore an explicit initialization strategy that uses the historical average from the previous year.

The results reveal distinct and meaningful functional roles across different dimensions. The historical dimension contributes more prominently to extreme event capture, and its benefit becomes more pronounced as the hierarchical receptive field expands. The temporal dimension, in contrast, uses historical information as an anchor and contributes more to prediction accuracy. The variable dimension models inter-variable dependencies, improving both accuracy and extreme event capture, while shuffling the variable dimension further enhances model performance.

Notably, joint modeling of the temporal and historical dimensions introduces a trade-off between prediction accuracy and extreme event capture. Temporal modeling tends to produce more accurate but smoother predictions, whereas historical modeling sacrifices part of the accuracy to better preserve extreme values. Incorporating both dimensions enables the model to achieve a balanced performance across the two objectives. In our analysis of the historical dimension Sec. \ref{hist_abl}, we further show that this trade-off can be continuously adjusted by controlling the historical scope, allowing the model to adapt to different forecasting scenarios.

Finally, when historical averaging is used for prediction initialization, the role of the historical dimension is largely weakened, leading to overly smoothed predictions. When the same initialization strategy is applied to baseline methods such as iTransformer, it fails to bridge long-term historical evolution with short-term temporal dynamics, causing model failure. This further highlights the essential capability of our design in integrating short-long-term weather dynamics for GSWF.

\subsubsection{Historical Dimension Analysis}

\begin{figure}[!t]
  \centering
  \includegraphics[width=\linewidth]{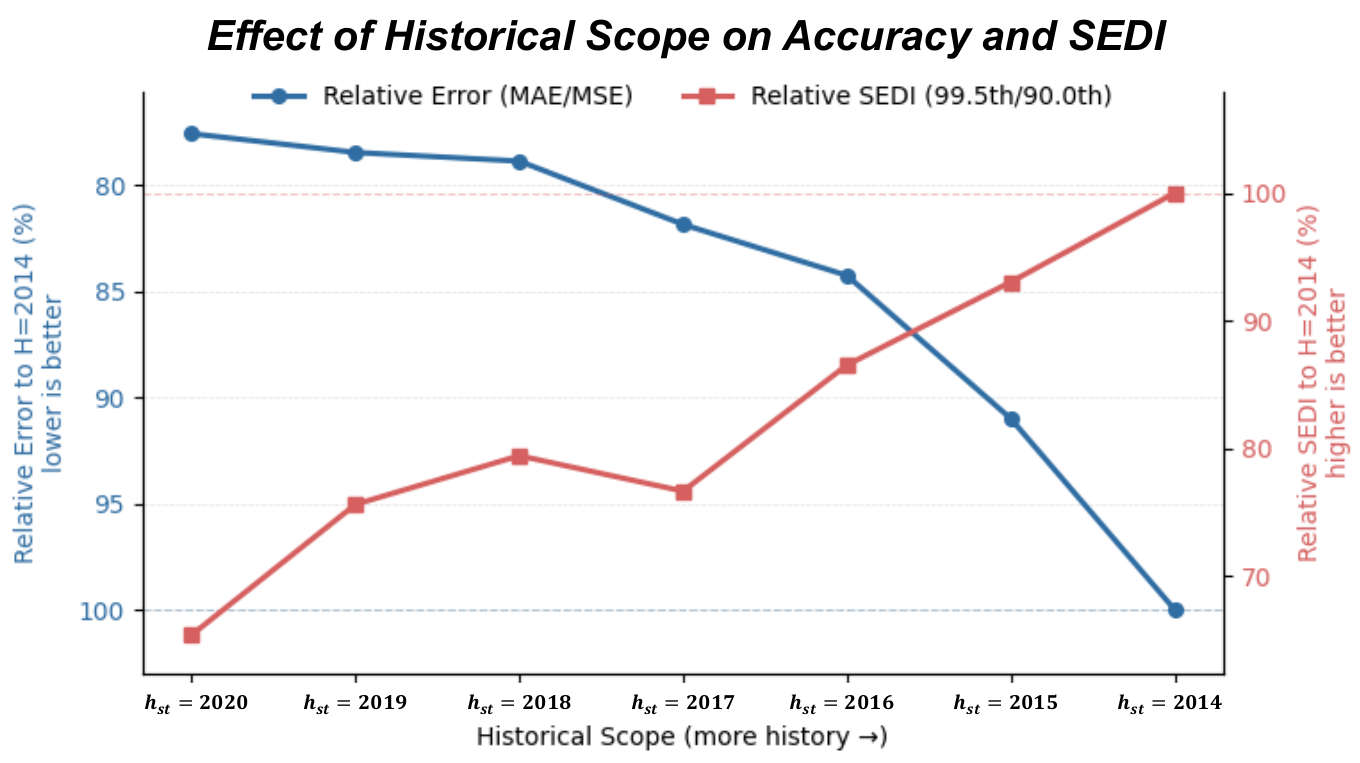}
  \caption{Effect of historical scope. Both curves are computed relative to the full historical setting starts from 2014 corresponds to 100\%. As the historical scope expands, TSSM achieves a controllable transition between accuracy-oriented forecasting and history-enhanced extreme event capture.}
  \label{fig:vis_hist}
\end{figure}

\label{hist_abl}
By expanding the historical dimension, we effectively improve GSWF performance. In the structural ablation study, the historical and temporal dimensions exhibit a clear division of labor, which motivates us to further analyze the historical dimension and investigate how this phenomenon can be regulated.

As shown in Fig. \ref{fig:vis_hist} and Tab. \ref{tab:main_result_hist}, we conduct an ablation analysis using historical training data from different starting years $h_{st}$. As the historical scope gradually expands, the model performance undergoes a continuous transition from last-year-anchored time-variable forecasting to history-enhanced forecasting. This gradual shift indicates that the trade-off between prediction accuracy and extreme event capture can be controlled by adjusting the scope of the historical dimension.

Furthermore, we investigate to what extent existing methods that do not explicitly model historical data can benefit from simply concatenating historical records into the temporal dimension, forming a longer look-back window.

As shown in Tab. \ref{tab:main_result_time}, we flatten eight years history into the time axis to construct a 384h look-back window. Most baseline methods achieve moderate performance improvements, but still significantly fall short of TSSM in effectively utilizing historical information. This demonstrates that transmitting historical information through the temporal dimension is far less efficient than modeling it with a dedicated historical dimension, further validating the effectiveness of our design.

\section{Limitation and Future Work}

As a preliminary attempt to extend GSWF along the historical dimension, this work adopts a simple and necessary design to incorporate period-aligned historical information. Although TSSM demonstrates clear effectiveness, we observe that historical information does not consistently improve all metrics, but instead introduces a trade-off between average accuracy and extreme event forecasting. Although continuous adjustment is achieved, replacing the fusion strategy between historical and temporal features with attention, gating, or direct interpolation does not fundamentally resolve this issue, nor does it produce continuous performance variations. These observations suggest that more effective and controllable historical-temporal fusion remains an important direction for future work.

\section{Conclusion}

In this work, we propose the Triaxial State Space Model, a history-enhanced temporal-variable-historical forecasting framework that extends GSWF beyond short-term temporal look-back windows. Inspired by climatological analysis in traditional meteorology, TSSM aligns station observations by the same month, day, and hour across years, reorganizing weather series into a time-variable-history structure. Based on this representation, we design hierarchically shared temporal, variable, and historical scanning modules to model short-term dynamics, inter-variable correlations, and long-term historical evolution. A causal forecasting framework further integrates current observations and historical references while maintaining training efficiency.

Comprehensive experiments verify the superiority of TSSM on Weather-5K and other human-involved datasets. TSSM achieves 10\% average accuracy gain and 61\% extreme event gain on Weather-5K, obtains 95\% best or second-best results on other datasets, and shows stronger advantages in long-term and robust forecasting. It reaches a 37.5\% overall gain at 240h and up to 103.5\% under iterative manner with cumulative iterative error $<$4.3\%, and exhibits strong robustness to missing data, with $<$ 10\% degradation compared with $>$ 50\% for baselines, at an 80\% missing rate. These results demonstrate that TSSM with historical-dimensional modeling provides a promising direction for more accurate, robust, and reliable global station weather forecasting.

\bibliographystyle{IEEEtran}
\bibliography{reference}

\newpage







\end{document}